\documentclass{article}

\usepackage{caption}
\usepackage{subcaption}
\usepackage[final]{neurips_2024}

\usepackage{algorithm}
\usepackage[noend]{algorithmic}
\usepackage[utf8]{inputenc} 
\usepackage[T1]{fontenc}    
\usepackage[dvipsnames]{xcolor}
\usepackage[colorlinks=true, allcolors=BlueViolet]{hyperref} 
\usepackage{graphicx}
\usepackage{url}            
\usepackage{booktabs}       
\usepackage{amsfonts}       
\usepackage{nicefrac}       
\usepackage{microtype}      
\usepackage{soul}
\usepackage{multirow} 

\usepackage{amsmath}  
\usepackage{xfrac}  
\usepackage{wrapfig}  
\usepackage{placeins}  

\newcommand{\myparagraph}[1]{\textbf{#1}  \hspace{0.4em}}

\title{QGFN: Controllable Greediness with Action Values}

\author{%
Elaine Lau$^{1\text{ }2\text{ }\footnote{dfasf}}$ \quad Stephen Zhewen Lu$^{2}$ \quad Ling Pan$^{1\text{ }5}$ \quad \\ \textbf{Doina Precup}$^{1\text{ }2\text{ }3}$ \quad \textbf{Emmanuel Bengio}$^4$ \\
$^1$Mila - Qu\'ebec AI Institute \quad $^2$McGill University \quad $^3$Google Deepmind \\
\quad $^4$Valence Labs \quad $^5$Hong Kong University of Science and Technology\\
}

\begin{document}

\maketitle

\begin{abstract}
Generative Flow Networks (GFlowNets; GFNs) are a family of energy-based generative methods for combinatorial objects, capable of generating diverse and high-utility samples. However, consistently biasing GFNs towards producing high-utility samples is non-trivial. In this work, we leverage connections between GFNs and reinforcement learning (RL) and propose to combine the GFN policy with an action-value estimate, $Q$, to create greedier sampling policies which can be controlled by a mixing parameter. We show that several variants of the proposed method, QGFN, are able to improve on the number of high-reward samples generated in a variety of tasks without sacrificing diversity.
\end{abstract}

\section{Introduction}

Generative Flow Networks~\citep{bengio2021flow, bengio2021gflownet}, also known as GFlowNets, or GFNs, were recently introduced as a novel generative framework in the family of energy-based models~\citep{malkin2022gflownets,zhang2022unifying}. Given some energy function $f(x)$ over objects $\mathcal{X}$, the promise of GFNs is to train a sampler $p_\theta$ such that at convergence $p_\theta(x) \propto \exp(-f(x))$; $\exp(-f(x))$ is also referred to as \emph{the reward} $R(x)$ in GFN literature, inheriting terminology from Reinforcement Learning (RL). GFNs achieve this sampling via a constructive approach, treating the creation of some object $x$ as a sequential additive process (rather than an iterative local process \textit{à la} Markov chain Monte Carlo (MCMC) that can suffer from mode-mixing issues). The main advantage of a GFN is its ability to generate a greater diversity of low-energy/high-reward objects compared to approaches based on MCMC or RL~\citep{bengio2021flow,jain2022biological,jain2023multi}, or even Soft-RL--which, while related to GFNs, accomplishes something different by default~\citep{tiapkin2023generative,mohammadpour2024maximum,deleu2024discrete}.

To generate more interesting samples and avoid  oversampling from low-reward regions, it is common to train a model to sample in proportion to $R(x)^\beta$; $\beta$ is an inverse temperature, typically $\gg 1$, which pushes the model to become \emph{greedier}. The use of this temperature (hyper)parameter is an important control knob in GFNs.  However, tuning this hyperparameter is non-trivial, which complicates training certain GFNs; for example, trajectory balance~\citep{malkin2022trajectory} is sensitive to the ``peakiness" of the reward landscape~\citep{madan2023learning}.
Although it is possible to train temperature-conditional models~\citep{kim2023learning}, doing so essentially requires learning a whole \emph{family} of GFNs--no easy task, albeit doable, e.g., in multiobjective settings~\citep{jain2023multi}.

In this work, we propose an approach that
allows selecting arbitrary greediness at inference time, which preserves the generality of temperature-conditionals, while simplifying the training process. We do so without the cost of learning a complicated family of functions and without conditionals, instead only training two models: 
a GFlowNet and an action-value function $Q$~\citep{watkins1992q,mnih2013playing}.

Armed with the forward policy of a GFN, $P_F$ (which decides a probability distribution over actions given the current state, i.e.~$\pi$ in RL), and the action-value, $Q$, we show that it is possible to create a variety of controllably greedy sampling policies, controlled by parameters that require no retraining. We show that it is possible to simultaneously learn $P_F$ and $Q$, and in doing so, to generate more high-reward yet diverse object sets.  
In particular, we introduce and benchmark three specific variants of our approach, which we call QGFN: $p$-greedy, $p$-quantile, and $p$-of-max. 

We evaluate the proposed methods on 5 standard tasks used in prior GFN works: the fragment-based molecular design task introduced by~\citet{bengio2021flow}, 2 RNA design tasks introduced by~\citet{sinai2020adalead}, a small molecule design task based on QM9~\citep{jain2023multi}, as well as a bit sequence task from~\citet{malkin2022trajectory,shen2023understanding}.
The proposed method outperforms strong baselines, achieving high average rewards and discovering modes more efficiently, sometimes by a large margin.
We conduct an analysis of the proposed methods, investigate key design choices, and probe the methods to understand why they work. We also  investigate other possible combinations of $Q$ and $P_F$--again, entirely possible at inference time, by combining trained $Q$ and $P_F$ models.

\section{Background and Related Work}
We follow the general setting of previous GFN literature and consider the generation of discrete finite objects, but in principle our method could be extended to the continuous case~\citep{lahlou2023theory}.

\textbf{GFlowNets} \hspace{0.4em} GFNs~\citep{bengio2021gflownet} sample objects by decomposing their generation process in a sequence $\tau = (s_0, .., s_T)$ of constructive steps. The space can be described by a pointed directed acyclic graph (DAG) $\mathcal{G} = (\mathcal{S}, \mathcal{A})$, where $s\in \mathcal{S}$ is a partially constructed object, and $(s\to s')\in\mathcal{A}\subset \mathcal{S} \times \mathcal{S}$ is a valid additive step (e.g., adding a fragment to a molecule). $\mathcal{G}$ is rooted at a unique initial state $s_0$.

GFNs are trained by pushing a model to satisfy so-called \emph{balance} conditions of flow, whereby flows $F(s)$ going through states are conserved such that terminal states (corresponding to fully constructed objects) are sinks that absorb $R(s)$ (non-negative) units of flow, and intermediate states have as much flow coming into them (from parents) as flow coming out of them (to children). This can be described succinctly as follows, for any partial trajectory $(s_n, .., s_m)$:
\begin{equation}
    F(s_n)\prod_{i=n}^{m-1} P_F(s_{i+1}|s_i) = F(s_m) \prod_{i=n}^{m-1} P_B(s_i |s_{i+1})
\end{equation}

where $P_F$ and $P_B$, the forward and backward policies, are distributions over children and parents respectively, representing the fraction of flow emanating forward and backward from a state. By construction for terminal (leaf) states $F(s)=R(s)$.

Balance conditions lead to learning objectives such as Trajectory Balance~\citep[TB;][]{malkin2022trajectory}, where $n=0$ and $m$ is the trajectory length, and Sub-trajectory Balance~\citep[SubTB;][]{madan2023learning}, where all combinations of $(n,m)$ are used. 
While a variety of GFN objectives exist, we use these two as they are considered standard. If those objectives are fully satisfied, i.e. 0-loss everywhere,  terminal states are guaranteed to be sampled with probability $\propto R(s)$~\citep{bengio2021flow}.

\textbf{Action values} \hspace{0.4em} For a broad overview of RL, see \citet{sutton2018reinforcement}. A central object in RL is the \emph{action-value} function $Q^\pi(s, a)$, which estimates the expected ``reward-to-go'' when following a policy $\pi$ starting in some state $s$ and taking action $a$; for some discount $0\leq \gamma\leq 1$,
\begin{equation}
    Q^\pi(s,a) = \mathbb{E}_{\substack{a_t\sim\pi(.|s_t) \\ \!\!\! s_{t+1}\sim T(s_t,a_t)}} \left[\sum^\infty_{t=0} \gamma^t R(s_t)\big|s_0=s, a_0=a\right]
\end{equation}
While $T(s,a)$ can be a stochastic transition operator, in a GFN context objects are constructed in a deterministic way (although there are stochastic GFN extensions;~\citet{pan2023stochastic}). 

Because rewards are only available for finished objects, $R(s)=0$ unless $s$ is a terminal state, and we use $\gamma=1$ to avoid arbitrarily penalizing ``larger'' objects.  Finally, as there are several possible choices for $\pi$, we will simply refer to $Q^\pi$ as $Q$ when statements apply to a large number of such choices.

\subsection{Related Work} 
\textbf{RL and GFlowNets} \hspace{0.4em} There are clear connections between the GFN framework and RL framework~\citep{tiapkin2023generative,mohammadpour2024maximum,deleu2024discrete}. Notably, \citet{tiapkin2023generative} show that it is possible to reformulate fixed-$P_B$ GFlowNets as a Soft-RL problem within a specific class of reward-modified MDPs. While they show that this reformulated problem can then be tackled with any Soft-RL method, this still essentially solves the original GFlowNet problem, i.e. learn $p_\theta(x) \propto R(x)$. Instead, we are interested in greedier-yet-diverse methods.

A commonly used tool in GFNs (and QGFN) to increase the average reward, aka ``greediness" of the drawn samples, is to adapt the reward distribution by using an altered reward function $\hat{R}(x) = R(x)^\beta$ and adjusting the exponent parameter $\beta$: the higher the $\beta$, the greedier the model should be~\citep{jain2023multi}. However, increasing $\beta$ often induces greater numerical instability (even on a log scale), 
and reduces diversity because the model is less incentivized to explore ``middle-reward'' regions. This can lead to mode collapse. \citet{kim2023learning} show that it is possible to train models that are conditioned on $\beta$, which somewhat alleviates these issues, but at the cost of training a more complex model. 

Again, while we could leverage the equivalence between the GFN framework and the Soft-RL framework~\citep{tiapkin2023generative}, this approach would produce a soft policy. We propose a different approach that increases greediness of the policy via ``Hard-RL''.

\textbf{Improving GFlowNet sampling} \hspace{0.4em} A number of works have also made contributions towards improving utility in GFlowNets, via local search~\citep{kim2023local}, utilizing intermediate signals~\citep{pan2023better}, or favoring high-potential-utility intermediate states~\citep{shen2023understanding}, as well as the use of RL tools such as replay buffers~\citep{vemgal2023empirical}, target networks~\citep{lau2023dgfn}, or Thompson sampling~\citep{rector2023thompson}.

\vspace{-0.25em}
\section{Motivation}
\vspace{-0.25em}
\begin{figure}
\vspace{-0.7cm}
\begin{center}
\centerline{\includegraphics[height=6cm]{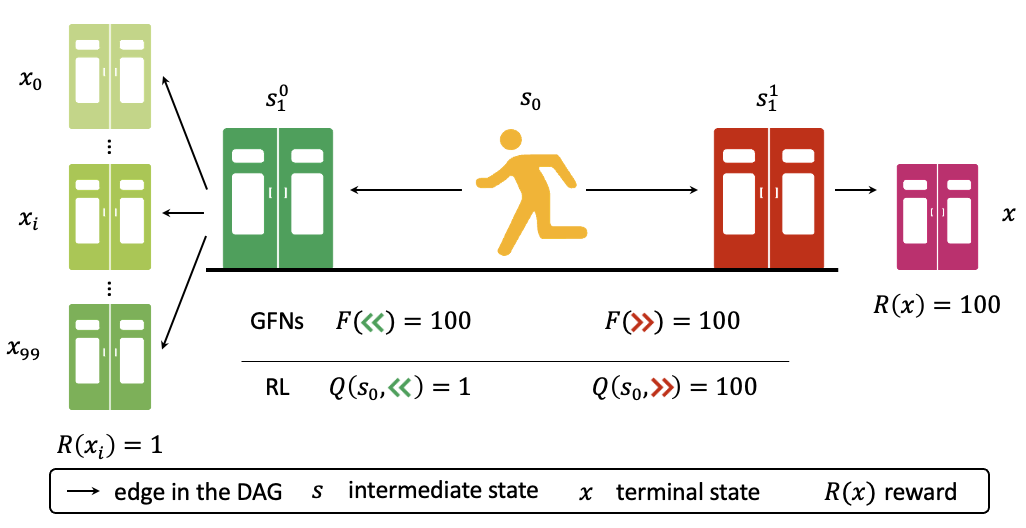}}
\caption{Solely relying on flow functions $F$ in GFNs can be insufficient. While GFNs capture \emph{how much stuff} there is, they spend time sampling from lots of small rewards.}
\label{fig:mot}
\end{center}
\vskip -0.23in
\end{figure}

Consider the following scenario, illustrated in Figure~\ref{fig:mot}: an agent is faced with two doors. Behind the left door, there are 100 other doors, each hiding a reward of 1. Behind the right door, there is a single door hiding a reward of 100. The flow will be such that $F(\mathrm{left})=F(\mathrm{right})=100$, meaning that a GFN agent will pick either door with probability $\sfrac{1}{2}$. The action value function is $Q(s_0, \mathrm{left})=1$, $Q(s_0, \mathrm{right})=100$, so an agent basing its decisions on $Q$ will reach for the door with reward 100.

This example shows that {\em relying solely on flows is not always sufficient to provide high-value samples frequently} and is illustrative of real-world scenarios. Consider molecular design (a very large search space) with some reward in $[0,1]$; there may be $10^6$ molecules with reward $.9$, but just a dozen with reward $1$. Since $.9\times 10^6$ is much bigger than $12 \times 1$, the probability of sampling a reward 1 molecule will be low if one uses this reward naively. While using a temperature parameter is a useful way to increase the probability of the reward $1$ molecules, we propose a complementary, inference-time-adjustable method.

\setulcolor{black}
\setul{0.3ex}{0.1ex}
Note that relying solely on $Q$ is also insufficient. If $Q$ were estimated very well for the optimal policy (which is extremely hard), it would be (somewhat) easy to find the reward 1 molecules via some kind of tree search following $Q$ values. However, in practice, RL algorithms easily collapse to non-diverse solutions, only discovering a few high reward outcomes. This is where flows are useful: because they capture \textit{how \ul{much} stuff} there is in a particular branch (rather than an expectation), it is useful to follow flows to find regions where there is potential for reward. In this paper, we propose a method that can be greedier (by following $Q$) while still being exploratory and diverse (by following $F$ through $P_F$).

\section{QGFN: controllable greediness through $Q$}
\label{sec:qgfn-method}

Leveraging the intuition from the example above, we now propose and investigate several ways in which we can use $Q$-functions to achieve our goal; we call this general idea \textbf{QGFN}. In particular, we present three variants of this idea, which are easy to implement and effective: $p$-greedy QGFNs, $p$-quantile QGFNs, and $p$-of-max QGFNs. In \S\ref{sec:results} and \S\ref{sec:analysis}, we show that these approaches provide a favourable trade-off between reward and diversity, during both training and  inference.

As is common in GFlowNets, we train QGFN by sampling data from some behavior policy $\mu$. We train $F$ and $P_F$ (and use a uniform $P_B$) to minimize a flow balance loss on the minibatch of sampled data, using a temperature parameter $\beta$. Additionally, we train a $Q$-network to predict action values on the same minibatch (the choice of loss will be detailed later). Training the GFN and $Q$ on a variety of behaviors $\mu$ is possible because both are off-policy methods. Indeed, instead of choosing $\mu$ to be a noisy version of $P_F$ as is usual for GFNs, we combine the predictions of $P_F$ and $Q$ to form a \emph{greedier} behavior policy. In all proposed variants, this combination is modulated by a factor $p\in[0,1]$, where $p=0$ means that $\mu$ depends only on $P_F$, and $p=1$ means $\mu$ is greediest, as reflected by $Q$. The variants differ in the details of this combination.

\myparagraph{$p$-greedy QGFN} Here, we define $\mu$ as a mixture between $P_F$ and the $Q$-greedy policy, controlled by factor $p$:
\begin{equation}
    \mu(s'|s) = (1-p)P_F(s'|s) + p\mathbb{I}[s' = \mathrm{argmax}_i Q(s, i)]
\end{equation}

In other words, we follow $P_F$, but with probability $p$, the greedy action according to $Q$ is picked. All states reachable by $P_F$ are still reachable by $\mu$. Note that $p$ can be changed to produce very different $\mu$ without having to retrain anything.

\myparagraph{$p$-quantile QGFN} Here, we define $\mu$ as a masked version of $P_F$, where actions below the $p$-quantile of $Q$, denoted $q_p(Q, s)$,  have probability $0$ (so are discarded):
\begin{equation}
    \mu(s'|s) \propto P_F(s'|s) \mathbb{I}[Q(s, s') \geq q_p(Q,s) ]
\end{equation}

This can be implemented by sorting $Q$ and masking the logits of $P_F$ accordingly. This method is more aggressive, since it prunes the search space, potentially making some states unreachable. 
Again, $p$ is changeable.

\myparagraph{$p$-of-max QGFN} Here, we define $\mu$ as a masked version of $P_F$, where actions with $Q$-values less than $p \max_a Q(s,a)$ have probability 0:
\begin{equation}
    \mu(s'|s) \propto P_F(s'|s) \mathbb{I}[Q(s, s') \geq p \max_i Q(s, i) ]
\end{equation}

This is similar to $p$-quantile pruning, but the number of pruned actions  changes as a function of $Q$. If all actions are estimated as good enough, it may be that no action is pruned, and vice versa, only the best action may be retained is none of the others are good.

This method also prunes the search space, and $p$ remains changeable. Note that in a GFN context, rewards are strictly positive, so $Q$ is also positive.

\myparagraph{Policy evaluation, or optimal control?}  In the design of the proposed method, we were faced with an interesting choice: what policy $\pi$ should $Q^\pi$ evaluate? The first obvious choice is to perform $Q$-learning~\citep{mnih2013playing}, and estimate the optimal value function $Q^*$, with a 1-step TD objective. As we detail in \S\ref{sec:analysis}, this proved to be fairly hard, and 1-step $Q_\theta$ ended up being a poor approximation. 

A commonly used trick to improve the performance of bootstrapping algorithms is to use $n$-step returns~\citep{hessel2018rainbow}. This proved essential to our work, and also revealed something curious: we've consistently found that, while results started improving at $n\geq 5$, a consistently good value of $n$ was close to the maximum trajectory length. This has an interesting interpretation, as beyond a certain value of $n$, $Q_\theta$ becomes closer to $Q^\mu$ and further from $Q^*$. In other words, using an ``on-policy'' estimate $Q^\mu$ rather than an estimate of the optimal policy seems beneficial, or at least easier to learn as a whole. In hindsight, this makes sense because on-policy evaluation is easier than learning $Q^*$, and since we are combining the $Q$-values with $P_F$, any method which biases $\mu$ correctly towards better actions is sufficient (we do not need to know exactly the best action, or its exact value).  

\myparagraph{Selecting greediness} In the methods proposed above, $p$ can be changed arbitrarily. We first note that we train with a constant or annealed\footnote{Specifically for $p$-quantile QGFN and $p$-of-max, we found training to be more stable if we started with $p=0$ and annealed $p$ towards its final value with a single \sfrac{1}{2}-period cosine schedule over 1500 steps.} value of $p$ and treat it as a standard hyperparameter in all the results reported in \S\ref{sec:results}. 

Second, as discussed in \S\ref{sec:analysis}, after training, $p$ can be changed with a predictable effect: the closer $p$ is to 1, the greedier $\mu$ becomes.
Presumably, this is because the model generalizes, and $Q$-value estimates for ``off-policy'' actions are still a reliable guess of the reward obtainable down some particular branch. When making $p$ higher, $Q$ may remain a good \emph{lower bound} of the expected reward (after all, $\mu$ is becoming greedier), which is still helpful.

Generally, such a policy will have reasonable outcomes, regardless of the specific $\mu$ and $p$ used during training.
Finally, it may be possible and desirable to use more complex schedules for $p$, or to sample $p$ during training from some (adaptive) distribution, but we leave this for future work.

\begin{figure}
\vskip 0.2in
\begin{center}
\centerline{\includegraphics[width=\textwidth]{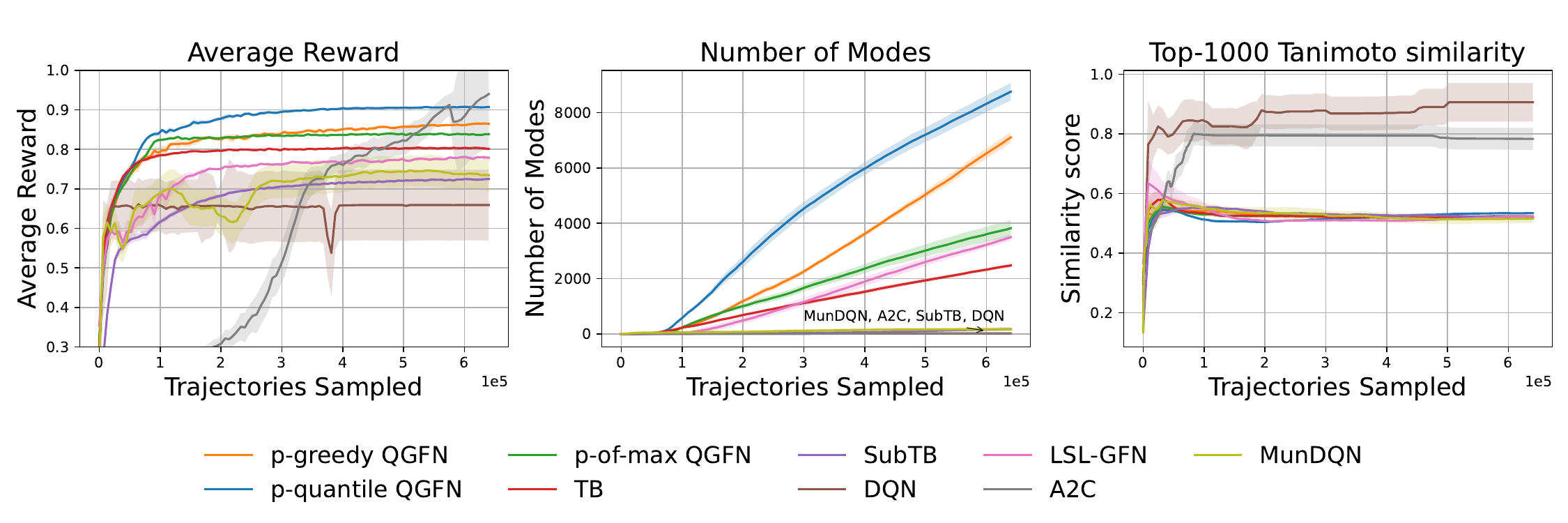}}
\vspace*{-1.5mm}
\caption{Fragment-based molecule task. \textit{Left:} Average rewards over the training trajectories. \textit{Center:} Number of unique modes with a reward threshold exceeding 0.97 and pairwise Tanimoto similarity score less than 0.65. \textit{Right:} Average pairwise Tanimoto similarity score for the top 1000 molecules sampled by reward. Lines are the interquartile mean and standard error calculated over 5 seeds.}
\label{fig:seh-results}
\end{center}
\vskip -0.4in
\end{figure}

\section{Main results}
\label{sec:results}

\begin{wrapfigure}{L}{7cm}
\vspace{-26pt}
\begin{center}
\centerline{\includegraphics[scale=0.3]{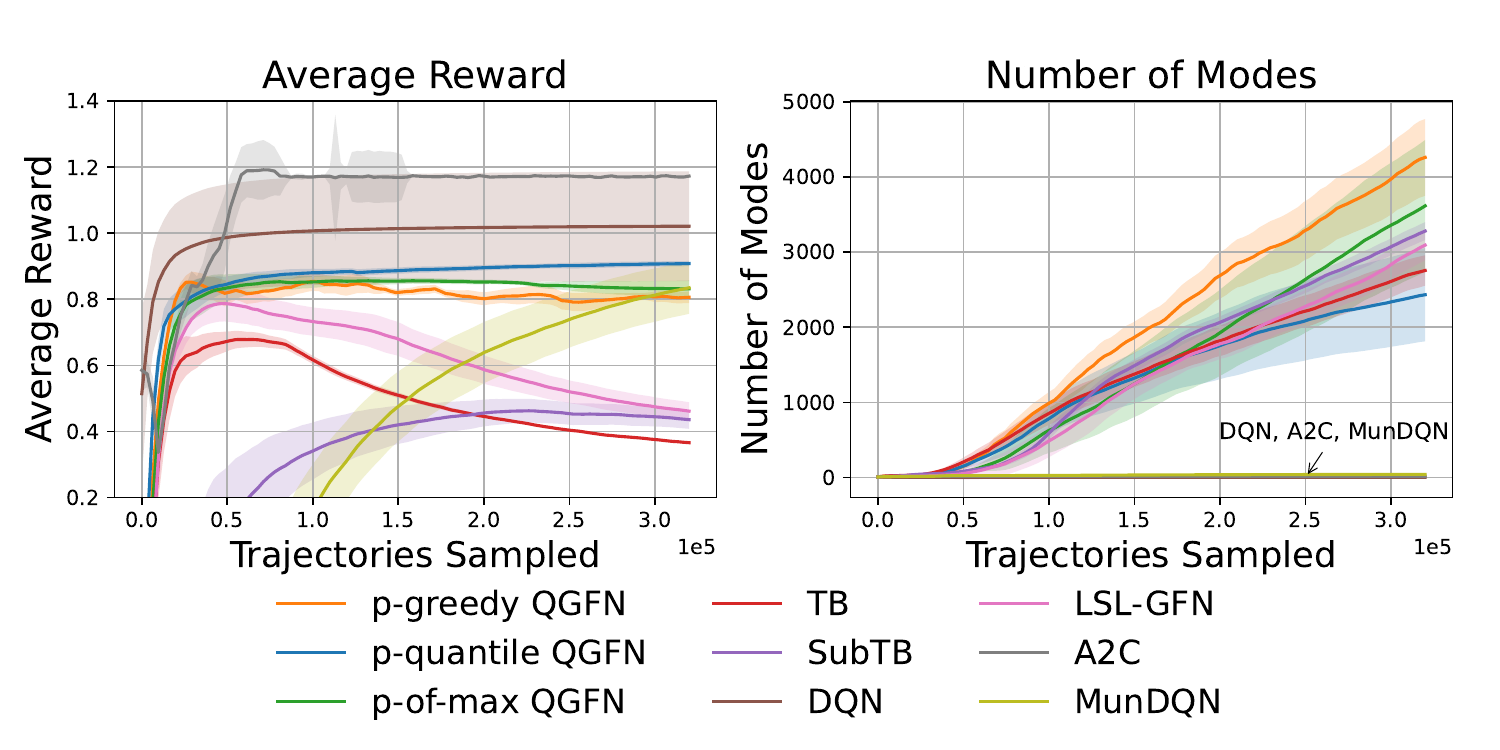}}
\caption{QM9 task. \textit{Left:} Average rewards over training trajectories. \textit{Right:} Number of modes with a reward above 1.10 and pairwise Tanimoto similarity less than 0.70.}

\label{fig:qm9-results}
\end{center}
\vspace{-0.9cm}
\end{wrapfigure}

We experiment on 5 standard tasks used in prior GFlowNet literature. As baselines, we use Trajectory Balance, Sub-Trajectory Balance, LSL-GFN~\citep{kim2023learning} i.e. \emph{learning to scale logits} which controls greediness through temperature-conditioning, and as RL baselines A2C~\citep{mnih2016asynchronous}, DQN~\citep{mnih2013playing} (which on its own systematically underperforms in these tasks), and \citet{tiapkin2023generative}'s MunDQN/GFlowNet.

We report the average reward obtained by the agents, as well as the total number of modes of the distribution of interest found during training. By \emph{mode}, we mean a high-reward object that is separated from previously found modes by some distance threshold. The distance function and threshold we use, as well as the minimum reward threshold for an object to be considered a mode, depend on the task.

\textbf{Fragment-based molecule generation task:}\footnote{We note that there exist a number of differing implementations of this task in the GFlowNet literature, we use that of \url{https://github.com/recursionpharma/gflownet}} Generate a graph of up to 9 fragments, where the reward is based on a prediction of the binding affinity to the sEH protein, using a model provided by \citet{bengio2021flow}. $|\mathcal{X}| \approx 10^{100}$, there are 72 available fragments, some with many possible attachment points. We use Tanimoto similarity~\citep{tanimoto2004}, with a threshold of 0.65, and a reward threshold of 0.97. Results are shown in Fig.~\ref{fig:seh-results}.

\textbf{Atom Based QM9 task:} Generate small molecules of up to 9 atoms following the QM9 dataset~\citep{qm9dataset2014}. $|\mathcal{X}| \approx 10^{12}$, the action space includes adding atoms or bonds, setting node or bond properties and stop. A MXMNet proxy model~\citep{MXMNET2020}, trained on QM9, predicts the HOMO-LUMO gap, a key indicator of molecular properties including stability and reactivity, and is used as the reward. Rewards are in the $[0, 2]$ range, with a $1.10$ threshold and a minimum Tanimoto similarity of $0.70$ to define modes. Results are shown in Fig.~\ref{fig:qm9-results}. 

\begin{figure}
\begin{center}

\centerline{\includegraphics[width=\textwidth]{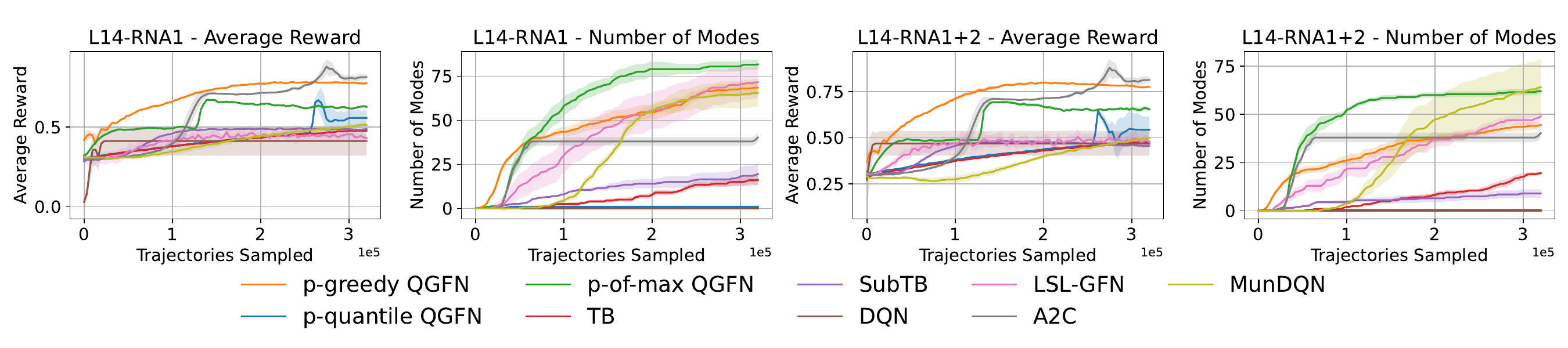}}
\caption{RNA-binding tasks, Average reward and modes. \textit{Left}:  L14RNA1 task. \textit{Right}: L14RNA1+2 task, based on 5 seeds (interquartile mean and standard error shown).}
\label{fig:rna-results}
\end{center}
\vskip -0.4in
\end{figure}

\begin{wrapfigure}{R}{7cm}
\vspace{-0.8cm}
\begin{center}
\centerline{\includegraphics[trim={0 0 0 1mm},clip,height=4.7cm]{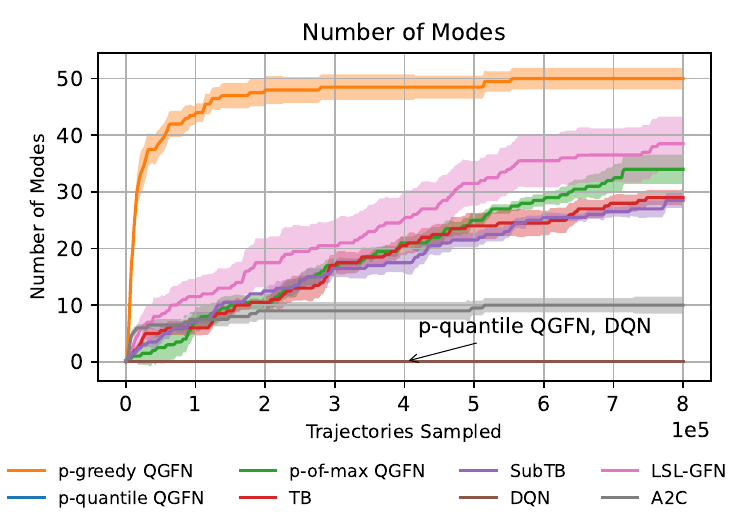}}
\caption{Bit sequence task, $k=1$. Interquartile mean and standard error over 5 seeds.}
\label{fig:bit-results}
\end{center}
\vskip -0.3in
\end{wrapfigure}

\textbf{RNA-binding task:} Generate a string of 14 nucleobases. The reward is a predicted binding affinity to the target transcription factor, provided by the ViennaRNA~\citep{lorenz2011viennarna} package for the binding landscapes; we experiment with two RNA binding tasks; L14-RNA1, and L14-RNA1+2 (two binding targets) with optima computed from~\citet{sinai2020adalead}. $|\mathcal{X}|$ is $4^{14}\approx 10^9$, there are 4 tokens: adenine (A), cytosine (C), guanine (G), uracil (U).  Results are shown in Fig.~\ref{fig:rna-results}.

\textbf{Prepend-Append bit sequences:} Generate a bit sequence of length 120 in a prepend-append MDP, where $|\mathcal{X}|$, limited to $\{0,1\}^n$, is $2^{120} \approx 10^{36}$. 
For a sequence of length $n$, $R(x)=\exp \left(1-\min _{y \in M} d(x, y) / n\right)$. A sequence is considered a mode if it is within edit distance $\delta$ from $M$, where $M$ is defined as per~\citet{malkin2022trajectory} (although the task we consider here is a more complex version, introduced by~\citet{shen2023understanding}, since prepend actions induce a DAG instead of a simpler tree). In our experiment, $|M|=60, n = 120, k=1, \delta = 28$, where $k$ is the bit width of actions. 
Results are shown in Fig.~\ref{fig:bit-results}.

\subsection{Analysis of results}

Across tasks, QGFN variants produce high rewards \emph{and} find a higher number of modes, i.e. high-reward dissimilar objects. The latter could seem surprising, because a priori, increasing the greediness of a method likely reduces its diversity. This  fundamental trade-off is known in RL as the exploration-exploitation dilemma~\citep{sutton1988learning,sutton2018reinforcement}. However, we are leveraging two methods and combining their strengths to reap the best from both worlds: GFlowNets are able to \emph{cover} the state space, because they  attempt to model all of it, by learning $p_\theta(x) \propto R(x)$, while $Q$ approximates the expected reward of a particular action, which can help guide the agent by selecting high-expected-reward branches. Another way to think about this: GFNs are able to estimate \emph{how many} high-reward objects there are in different parts of the state space. The agent thus ends up  going in all important regions of the state space, but by being a bit more greedy through $Q$, it focuses on higher reward objects, so it is more likely to find objects with reward past the mode threshold. To further understand the performance of QGFN, we formally analyse a bandit setting, and include derivations to illustrate the general case, in Appendix \S\ref{sec:app-derivation}.

We also report the average reward and pairwise similarity for the fragment task based on 1000 samples over 5 seeds taken after training in Table \ref{table:inference-r-and-sim}. Again, QGFNs outperform GFNs in reward, while retaining low levels of inter-sample similarity. We again note that  at inference, we are able to use a different (and better) $p$ value than the one used at training time. We expand on this in \S\ref{sec:analysis}, and show that it is easy to tune $p$ to achieve different reward-diversity trade-offs at inference. The exact $p$ values used for Table~\ref{table:inference-r-and-sim} are provided in Appendix \S\ref{sec:experiment_details}.  Also note that in LSL-GFN $\beta$ is tunable at inference, and in Table \ref{table:inference-r-and-sim} we choose the $\beta$ value such that average similarity is near the 0.65 mode threshold we use (choosing a greedier $\beta$ induces a collapse in diversity).

\begin{table}[t]
\caption{Fragment-based molecule task: Reward and Diversity at inference after training. }
\label{table:inference-r-and-sim}
\vspace{-0.05cm}
\begin{center}
\begin{small}
\begin{sc}
    \centering

\begin{tabular}{@{\hspace{3pt}}lll@{\hspace{2pt}}}
\toprule
Method           & Reward($\uparrow$) & Similarity ($\downarrow$) \\ \hline
GFN-TB                    &    0.780$\pm$0.003    &    0.545$\pm$0.002       \\
GFN-SubTB                   &    0.716$\pm$0.006     &      0.513$\pm$0.003     \\
LSL-GFN                     &   0.717$\pm$0.020        &    0.689$\pm$0.062     \\
p-greedy QGFN            &    0.950$\pm$0.004    &      0.551$\pm$0.015    \\
p-of-max \makebox[0pt][l]{QGFN}           &   \textbf{0.969$\pm$0.003}    &      0.514$\pm$0.001     \\
p-quantile \makebox[25pt][l]{QGFN}&     0.955$\pm$0.003   &     \textbf{0.509$\pm$0.008}     \\
\bottomrule
\end{tabular}

\end{sc}
\end{small}
\end{center}
\vskip -0.1in
\end{table}

\myparagraph{QGFN variants matter} We point the reader to an interesting result, which is consistent with our understanding of the proposed method. In the fragment task, the number of actions available to the agent is quite large, ranging from about 100 to 1000 actions depending on the state, and the best performing QGFN variant is one that consistently masks most actions: $p$-quantile QGFN. It is likely indeed that most actions are harmful, as combining two fragments that do not go together may be irreversibly bad, and masking helps the agent avoid undesirable regions of the state space. However, masking a fixed ratio of actions can provide more stable  training.

On the other hand, in the RNA design task, there are only 5 actions (4 nucleotides \texttt{ACGU} \& stop). We find that masking a constant \emph{number} of actions is harmful--it is likely that in some states all of them are relevant. So, in that task, $p$-greedy and $p$-of-max QGFN work best. 
This is also the case in the bit sequence task, for the same reasons (see Fig.~\ref{fig:bit-results}). To confirm this, we repeat the bit sequence task but with an expanded action space consisting not just of $\{0,1\}$, but of all 16 ($2^4$) sequences of 4 bits, i.e. $\{0000, 0001, .., 1111\}$. We find, as shown in Fig \ref{fig:bit-results-4}, that $p$-quantile indeed no longer underperforms.

\section{Method Analysis}
\label{sec:analysis}

We now analyze the key design choices in QGFN. We start by investigating the impact of $n$ (the number of bootstrapping steps in $Q$-Learning) and $p$ (the mixture parameter) on \emph{training}. We then look at trained models, and reuse the learned $Q$ and $P_F$ to show that it is possible to use a variety of sampling strategies, and to change the mixture factor $p$ to obtain a spectrum of greediness at test time. Finally, we empirically probe models to provide evidence as to \emph{why} QGFN is helpful.

\begin{figure}[!t]
\centering
\begin{minipage}{.5\textwidth}
  \centering
  \includegraphics[width=\linewidth]{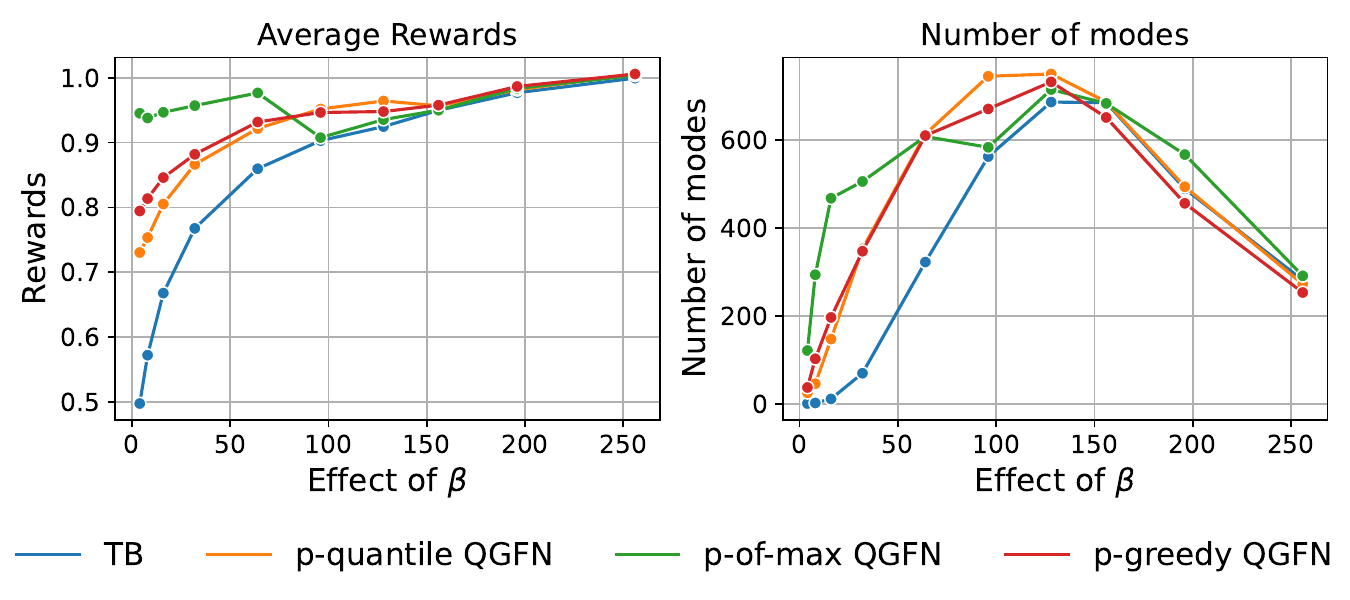}
\end{minipage}%
\begin{minipage}{.5\textwidth}
  \centering
  \includegraphics[width=\linewidth]{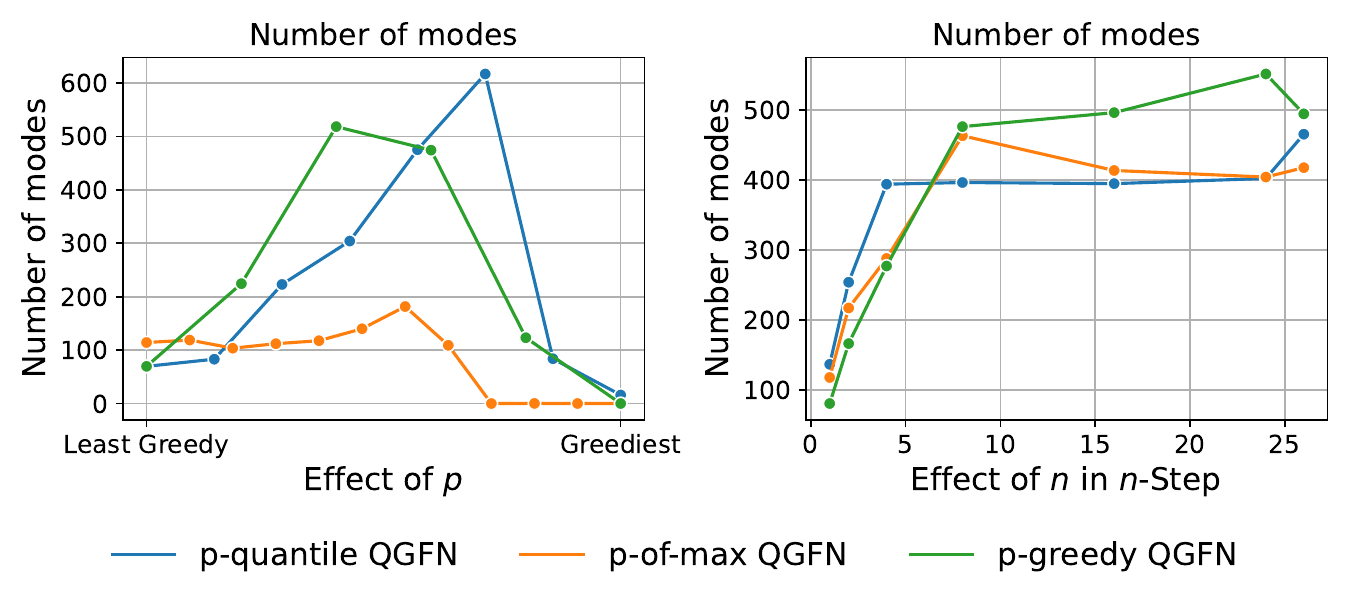}
\end{minipage}
\caption{Fragment task. \textit{Left}: Effect of $\beta$: Increasing greediness through $\beta$ increases the average reward but may lead to diversity collapse. QGFN maintains diversity with a lower $\beta$, while GFN collapses. Modes are counted from 1000 samples at inference, using an inference-adjusted $p$. \textit{Right}: Effect of training parameters $p$, and $n$: Changing $p$ can control greediness, while increasing $n$ is generally beneficial. Modes are counted from 1000 samples generated using the training $p$.}
\label{fig:hp-ablation}
\vskip -0.2in
\end{figure}

\myparagraph{Impact of $\beta$ in QGFN}  Fig.~\ref{fig:hp-ablation} shows the effect of training with different $\beta$ values on the average reward and number of modes when taking 1000 samples after training is done in the fragment task (over 5 seeds). As predicted, increasing $\beta$ increases the average reward of the agent, but at some point, causes it to become dramatically less diverse. As discussed earlier, this is typical of GFNs with a too high $\beta$, and is caused by a collapse around high-reward points and an inability for the model to further explore. While QGFN is also affected by this, it does not require as drastic values of $\beta$ to obtain a high average reward and discover a high number of modes.

\myparagraph{Impact of $n$ in QGFN} As mentioned in \S\ref{sec:qgfn-method}, training $Q$ with 1-step returns is ineffective and produces less useful approximations of the action value. Fig.~\ref{fig:hp-ablation} shows the number of modes within 1000 post-training samples in the fragment tasks, for models trained with a variety of $n$-step values. Models start being consistently good at $n=5$ and values  close to the maximum length of a trajectory tend to work well too.

\myparagraph{Impact of the training $p$ in QGFN} While our method allows changing $p$ more or less at will, we still require some value during training.  Fig.~\ref{fig:hp-ablation} shows that there are clear trade-offs between choices of $p$, some yielding significantly better diversity than others. For example, $p$-of-max is fairly sensitive to the chosen value during training, and for the fragment task doesn't seem to perform particularly well during training (especially when not annealed). On the other hand, as we will see in the next paragraph (and is also seen in Fig.~\ref{fig:hp-ablation}), $p$-of-max excels at inference, and is able to generate diverse and high-reward samples by adjusting $p$.

\myparagraph{Changing strategy after training} We now look at the impact of changing the mixture parameter $p$ and the sampling strategy for already trained models on average reward and average pairwise similarity. We use the parameters of a model trained with $p$-greedy QGFN, $p=0.4$. 

\begin{wrapfigure}{R}{7cm}
\vspace{-0.5cm}
\begin{center}
\centerline{\includegraphics[height=4.5cm]{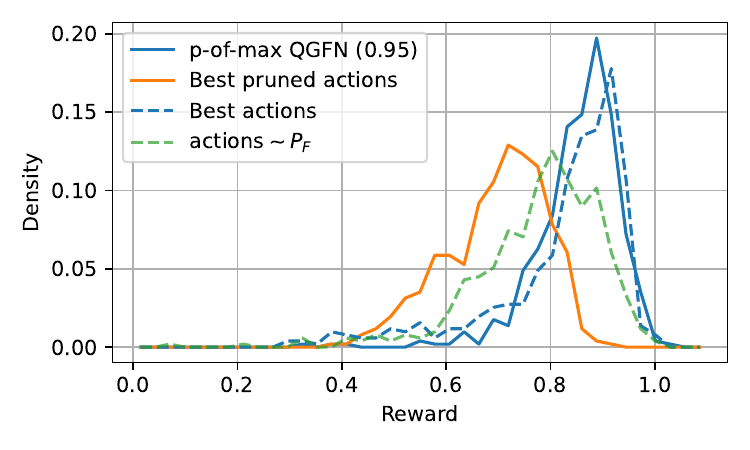}}
\caption{Pruning helps avoid low-reward parts of the state space. Reward distributions when (a) sampling with $p$-of-max; (b) greedily according to $P_F$ selecting actions that $p$-of-max would prune, \emph{Best pruned actions}; (c) selecting most likely $P_F$ actions regardless of $Q$, \emph{Best actions}; and (d) normal sampling from $P_F$ (without using $Q$).}
\label{fig:is_pruning_helpful}
\end{center}
\begin{center}
\vspace{-0.3cm}
\centerline{\includegraphics[height=4.5cm]{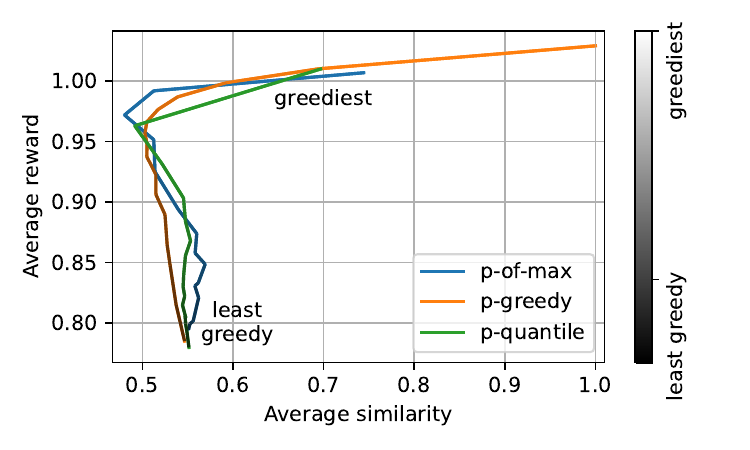}}
\caption{Varying $p$ at inference time induces reward-diversity trade-offs; fragment task.}
\label{fig:inference-p}
\end{center}
\vskip -0.5in
\end{wrapfigure}

With this model, we sample 512 new trajectories for a series of different $p$ values. For $p$-greedy and $p$-quantile, we vary $p$ between $0$ and $1$; for $p$-of-max, we vary $p$ between $.9$ and $1$ (values below $.9$ have minor effects). We visualize the effect of $p$ on reward and similarity statistics in Fig.~\ref{fig:inference-p}. 

First, we note that increasing $p$ has the effect we would hope, increasing the average reward. Second, we note that this works without any retraining; even though we (a) use values of $p$ different than those used during training, and (b) use QGFN variants different than those used during training, the behavior is consistent: $p$ controls greediness. Let us emphasize (b): even though we trained this $Q$ with $p$-greedy QGFN, we are able to use the $Q(s,a)$ predictions just fine with entirely different sampling strategies. This has some interesting implications; most importantly, it can be undesirable to \emph{train} with too high values of $p$ (because it may reduce the diversity to which the model is exposed), but what is learned transfers well to sampling new, high-reward objects with different values of $p$ and sampling strategies.

\begin{wrapfigure}{R}{7cm}
\vspace{-0.8cm}
\begin{center}
\centerline{\includegraphics[height=4.0cm]{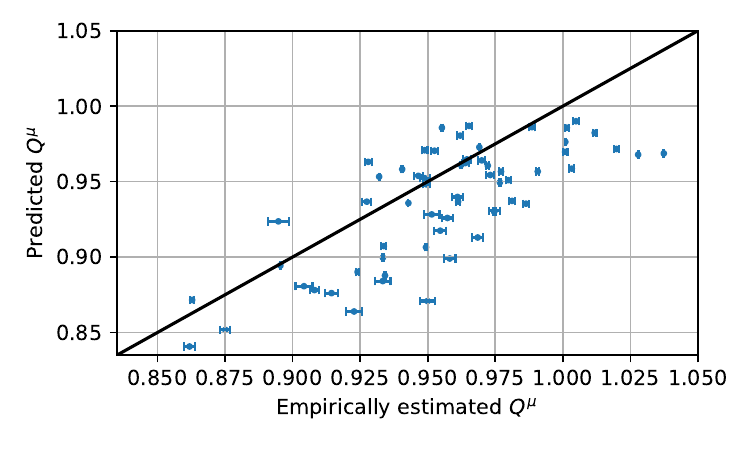}}
\caption{Comparing $Q(s,a; \theta)$ predictions with empirical estimates obtained by rollouts. Bars are standard error. $Q$ is relatively capable to estimate the returns of the corresponding policy.}
\label{fig:is_q_accurate}
\end{center}
\vskip -0.3in
\end{wrapfigure}

Finally, these results suggest that we should be able to prototype new QGFN variants, including expensive ones (e.g. MCTS) without having to retrain anything. We illustrate the performance of a few other variants in \S\ref{sec:extra-variants}, Fig.~\ref{fig:extra_variants}.

\myparagraph{Is $Q$ calibrated?} For our intuition of why QGFN works to really pan out, $Q$ has to be accurate enough to provide useful guidance towards high-reward objects. We verify that this is the case with the following experiment. We take a trained QGFN model ($p$-greedy, $p=0.4$, fragment task, maximum $n$-step) and sample $64$ trajectories. For each of those trajectories, we take a random state within the trajectory as a starting point, thereafter generating $512$ new trajectories. We then use the reward of those $512$ trajectories as an empirical estimate $\hat Q$ of the expected return, which $Q$ should roughly be predicting.  Fig.~\ref{fig:is_q_accurate} shows that this is indeed the case. Although $Q$ is not perfect, and appears to be underestimating $\hat{Q}$, it is making reasonable predictions.

\myparagraph{Is $Q$ really helpful?} In this experiment, we verify our intuition that pruning based on $Q$-values is helpful. We again take a trained model for the fragment task, and sample $512$ trajectories. We use $p$-of-max QGFN ($p=0.95$), and compare it to a strategy estimating \emph{best pruned actions}: for each trajectory, after some random number of steps $t\sim U[4,20]$ (the max is 27), we start deterministically selecting the action that is the most probable according to $P_F$ but would be masked according to $Q$. To ensure that this is a valid thing to do, we also simply look at \emph{Best actions}, i.e. after $t\sim U[4,20]$ steps,  deterministically select the action that is the most probable according to $P_F$, regardless of $Q$. 

Fig.~\ref{fig:is_pruning_helpful} shows that our sanity check, \emph{Best actions}, receives reasonable rewards, while selecting actions that would have been pruned leads to much lower rewards. The average likelihood from $P_F$ of these pruned actions was $.035$, while the average number of total actions was $\approx 382$ (and $1/382\approx 0.0026$). This confirms our hypothesis that $Q$ indeed masks actions that are likely according to $P_F$ but that do \textbf{not} consistently lead to high rewards.

\myparagraph{Why does changing $p$ work?} Recall that for QGFN to be successful, we rely on $n$-step TD, and therefore on somewhat ``on-policy'' estimates of $Q^\mu$. $\mu$ is really $\mu_p$, a function of $p$, meaning that if we change $p$, say to $p'$, during inference, $Q^{\mu_p}$ is not an accurate estimate of $Q^{\mu_{p'}}$. If this is the case, then there must be a reason why it is still helpful to prune based on $Q^{\mu_p}$ while using $\mu_{p'}$. In Fig.~\ref{fig:is_Q_helpful_when_changing_p},we perform the same measurement as in Fig.~\ref{fig:is_q_accurate}, but we change the $p$ value used to measure $\hat Q^{\mu_{p'}}$. We find that, while the rank correlation drastically goes down (although it stays well above 0), $Q^{\mu_p}$ remains helpful because it \emph{lower bounds} $\hat Q^{\mu_{p'}}$. 
If we prune based on $Q^{\mu_p}$, then we would want it to not lead us astray, and \emph{at least} make us greedier as we increase $p$. This means that if an action is not pruned, then we expect samples coming from it to be \emph{at least as good} as what $Q^{\mu_p}(s,a)$ predicts (in expectation). This is indeed the case.

Note that reducing $p$ simply leads $\mu$ to behave more like $P_F$, which is still a good sampler, and to rely less on $Q$, whose imperfections will then have less effect anyways.

\section{Conclusion and Discussion}
\label{sec:conclusion}
\vspace{-0.25em}
In this paper, we showed that by jointly training GFNs and $Q$-functions, we can combine their predictions to form behavior policies that are able to sample larger numbers of diverse and high-reward objects. These policies' mixture parameter $p$ is adjustable, even after training, to modulate the greediness of the resulting policy. We implement multiple ways of combining GFNs and $Q$s, referring to the general idea as QGFN: taking a greedy with probability $p$ ($p$-greedy QGFN), restricting the agent to the top $1-p$\% of actions ($p$-quantile QGFN), and restricting the agent to actions whose estimated value is at least a fraction $p$ of the best possible value ($p$-of-max QGFN).

We chose to show several variants of QGFN in this paper, because they all rely on the same principle, learning $Q$, but have different properties, which lead to better or worse behavior in different tasks. For example, pruning too aggressively on a task with a small number of actions is harmful. We also hope that by showing such a diversity of combinations of $P_F$ and $Q$, we encourage future work that combines GFNs and RL methods in novel and creative ways.

We also analyzed why our method works. We showed that the learned action-value $Q$ is predictive and helpful in avoiding actions that have high probability under $P_F$ but lower expected reward. Even when $Q$ predictions are not accurate, e.g. because we sample from a different policy than the one which $Q$ models, they provide a helpful lower bound that facilitates controllable greediness.

Our analysis suggests that at training time, QGFN works because it helps the agent to ``waste'' less time and capacity modeling low-reward objects, and that conversely the policy family that QGFN learns is able to sample more distinct high-reward objects given the same budget. In this sense, QGFN benefits from the advantages of both the GFlowNet and ``Hard``-RL frameworks.

\myparagraph{What didn't work} The initial stages of this project were quite different. Instead of combining RL and GFNs into one sampling policy, we instead trained two agents, a GFN and a DQN. Since both are off-policy methods we were hoping that sharing ``greedy'' DQN data with a GFN would be fine and make GFN better on high-reward trajectories. This was not the case, instead, the DQN agent simply slowed down the whole method--despite trying a wide variety of tricks, see \S\ref{sec:bad-dqgfn}.

\myparagraph{Limitations} Because we train two models, our method requires more memory and FLOPs, and consequently takes more time to train compared to TB (as shown in Table~\ref{tab:training_inference_time}). QGFN is also sensitive to how well $Q$ is learned, and as we've shown $n$-step returns are crucial for our method to work. In addition, although the problems we tackle are non-trivial, we do not explore the parameter and compute scaling behaviors of the benchmarked methods.

\myparagraph{Future work} We highlight two straightforward avenues of future work. First, there probably exist more interesting combinations of $Q$ and $P_F$ (and perhaps $F$), with different properties and benefits. Second, it may be interesting to further leverage the idea of pruning the action space based on $Q$, forming the basis for some sort of constrained combinatorial optimization. By using $Q$ to predict some expected property or constraint, rather than reward, we could prune some of the action space to avoid violating constraints, or to keep some other properties below some threshold (e.g. synthesizability or toxicity in molecules).

Finally, we hope that this work helps highlight the differences between RL and GFlowNet, while adding to the literature showing that these approaches complement each other well. It is likely that we are only scratching the surface of what is possible in combining these two frameworks.

\FloatBarrier
\newpage

\section*{Acknowledgements}

The bulk of this research was done at Valence Labs as part of an internship, using computational resources there. This research was also enabled in part by computational resources provided by Calcul Québec, Compute Canada and Mila. Academic authors are funded by their respective academic institution, Fonds Recherche Quebec through the FACSAcquity grant, the National Research Council of Canada and the DeepMind Fellowships Scholarship.

The authors are grateful to Yoshua Bengio, Moksh Jain, Minsu Kim, and the Valence Labs team for their feedback, discussions, and help with baselines.
\section*{Author Contributions} 
The majority of the experimental work, code, plotting, and scientific contributions were by EL, with support from EB. SL helped run some experiments, baselines and plots. The project was supervised by EB, and DP and LP provided additional scientific guidance. Most of the paper was written by EB. DP, LP, and EL contributed to editing the paper.

{\small \bibliography{neurips_2024}}
\bibliographystyle{plainnat}
\newpage 

\appendix 
\section{Analysing $p$-greedy}
\label{sec:app-derivation}

Consider the bandit setting where trajectories are 1 step and just consist in choosing a terminal state. Let $p_G(s) = R(s)/Z$. Let $0<p<1$, then with $\mu(s'|s) = (1-p)P_F(s'|s) + p\mathbb{I}[s'=\arg\max Q(s,s')]$, assuming there is only a single argmax $s^*$, then $p_\mu(s) = (1-p)R(s)/Z + p\mathbb{I}[s=\arg\max R(s)]$. This means that for every non-argmax state, $p_\mu(s) = (1-p) p_G(s) < p_G(s)$. We get that $\mathbb{E}_\mu[R] > \mathbb{E}_G[R]$:

\begin{align}
\mathbb{E}_\mu[R] - \mathbb{E}_G[R]=&\sum_s p_\mu(s) R(s) - \sum_s p_G(s) R(s)\\
=& (p + (1-p)R(s^*)/Z - R(s^*)/Z) R(s^*) + \sum_{s\neq s^*} (1-p)R(s)^2/Z-R(s)^2/Z\\
=& pR(s^*) - pR(s^*)^2/Z  + \sum_{s\neq s^*} (-p)R(s)^2/Z\\
=& p/Z \left(R(s^*)Z - R(s^*)^2  - \sum_{s\neq s^*} R(s)^2\right),\;\;\;Z=\sum_s R(s)\\
=& p/Z \left(R(s^*)[\sum_s R(s)] - R(s^*)^2  - \sum_{s\neq s^*} R(s)^2\right)\\
=& p/Z \left(R(s^*)[\sum_s R(s)] - \sum_{s} R(s)^2\right)\\
=& p/Z \left(\sum_{s} R(s^*)R(s) - R(s)^2\right)\\
\end{align}

since $R(s^*) > R(s)$ and both are positive then $R(s^*)R(s) > R(s)^2$ thus the last sum is positive. All other terms are positive, therefore $\mathbb{E}_\mu[R] - \mathbb{E}_G[R] > 0$.

In the more general case, we are not aware of a satisfying closed form, but consider the following exercise. 

Let $m(s,s') = \mathbb{I}[s'=\arg\max Q(s,s')]$. Let $F'$ be the "QGFN flow" which we'll decompose as $F'=F_G + F_Q$ where we think of $F_G$ and $F_Q$ as the GFN and Q-greedy contributions to the flows. Then:

\begin{align}
F'(s) &= \sum_{z\in\mathrm{Par}(s)} F'(z)((1-p)P_F(s|z) + pm(z,s))\\
&=\sum_z F'(z)(1-p)P_F(s|z) + \sum_z F'(z) pm(z,s)\\
&=\sum_z F_G(z)(1-p)P_F(s|z) + F_Q(z)(1-p)P_F(s|z)+ \sum_z F'(z) pm(z,s)\\
&=(1-p)F_G(s) + \sum_z F_Q(z)\mu(z|s)+ F_G(z) pm(z,s)
\end{align}

Recall that $p(s)\propto F(s)$. Intuitively then, the probability of being in a state is reduced by a factor $(1-p)$, but possibly increased by this extra flow that has two origins. First, flow $F_Q$ carried over by $\mu$, and second, new flow being "stolen" from $F_G$ from parents when $m(z, s)=1$, i.e. when $s$ is the argmax child. 

This suggests that flow (probability mass) in a $p$-greedy QGFN is smoothly redirected towards states with locally highest reward from ancestors of such states. Conversely, states which have many ancestors for which they are not the highest rewarding descendant will have their probability diminished.

\section{Additional experiments and analyses}

In this section, we provide additional experiments to support our main findings. We explore the use of $Q$ functions from different behavior policies, assess various QGFN inference variants, examine QGFN variants trained with alternative objectives, and investigate the effects of weight sharing in QGFN models.

\begin{figure}[h]
\vskip 0.2in
\begin{center}
\centerline{\includegraphics[width=0.5\columnwidth]{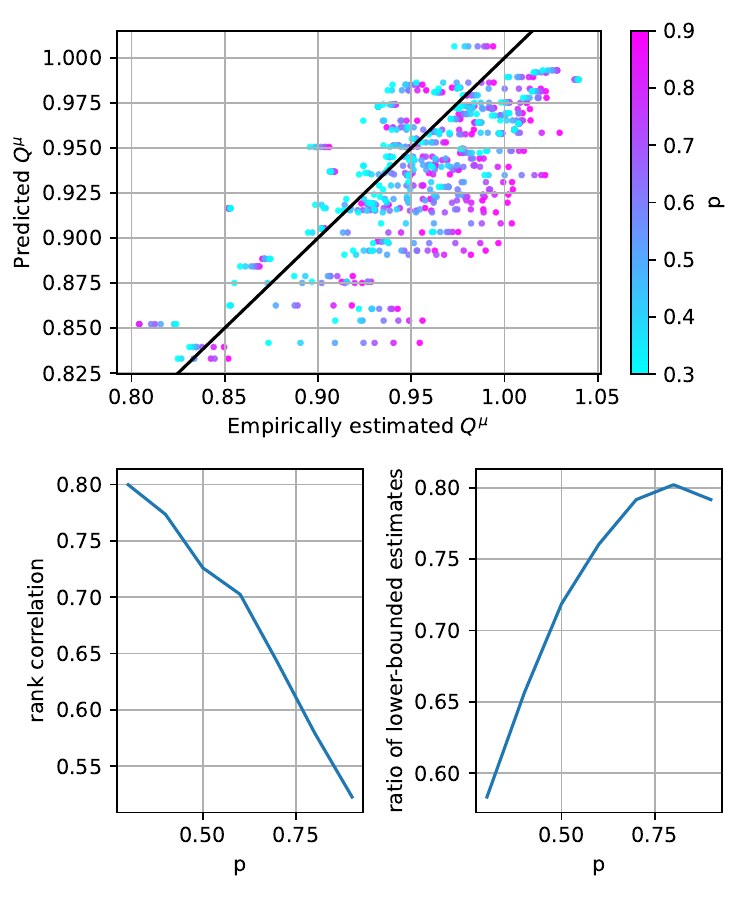}}
\caption{What happens to $Q(s,a)$ when changing $p$? We show here that while the rank correlation between $Q$ and the empirically estimated expected reward $\hat{Q}^{\mu_p}$ goes down when changing $p$, $Q$ remains a useful estimate in that it mostly lower bounds $\hat{Q}^\mu$. This means that, at worst, pruning based on the ``wrong'' $Q$ \& $p$ combination drops some high-reward objects, but does not introduce more lower-reward objects.}
\label{fig:is_Q_helpful_when_changing_p}
\end{center}
\vskip -0.2in
\end{figure}

\begin{figure}[h]
\vskip 0.2in
\begin{center}
\centerline{\includegraphics[width=0.5\columnwidth]{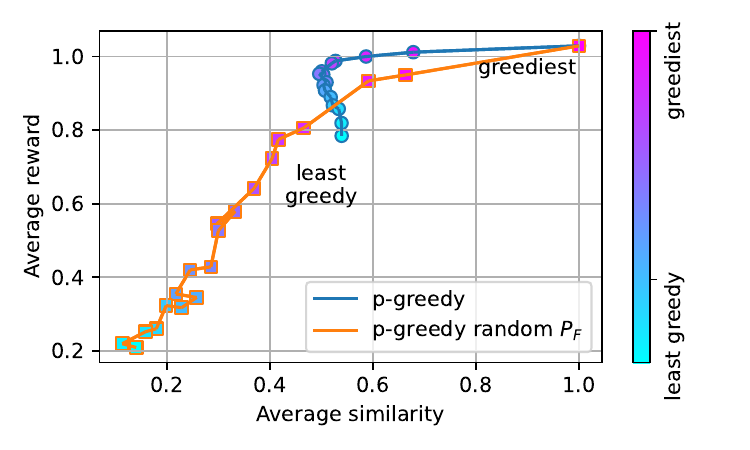}}
\caption{Comparison of trained $P_F$ vs. untrained $P_F$ with a trained $Q$ during inference on fragment-based molecule task}
\label{fig:inference_p_tradeoffs_v2}
\end{center}
\vskip -0.2in
\end{figure}

\subsection{Using $Q$ from different behavior policies}

Another approach we explored involves using a trained $Q$ function during inference that was trained on an entirely different behavior policy. Similarly, we apply the QGFN algorithm at each state of sampling trajectories during inference, but with a key difference: it is directly applied to a baseline model that has been trained independently. This approach aims to examine if a previously trained $Q$, when used in a different training setup but the same task, can guide independently trained models that may not perform as well during training but, with this assistance, can achieve significantly better results at inference. For instance, as shown in Figure~\ref{fig:seh-results}, the samples generated by SubTB average around 0.7 rewards throughout training. However, using the trained $Q$ as a greediness signal during inference allows us to discover samples with significantly higher rewards. Table~\ref{table:inference-w-other-objectives} details the effects of applying $Q$ during inference on independently trained baseline models for 1000 samples post-training of 5 seeds.

\begin{table}
\caption{Fragment-based molecule task: reward and diversity of independently trained baseline models using a trained $Q$. The $p$ values for $p$-greedy, $p$-of-max, and $p$-quantile QGFN are set at 0.4, 0.9858, and 0.93, respectively.}
\label{table:inference-w-other-objectives}
\vspace{0.2cm}
\centering
\begin{small}
\begin{sc}
\begin{tabular}{@{}lcccccc@{}}
\toprule
\multirow{2}{*}{Variant} & \multicolumn{2}{c}{TB} & & \multicolumn{2}{c}{SubTB} \\
\cmidrule{2-3} \cmidrule{5-6}
 & Reward & Diversity & & Reward & Diversity \\
\midrule
baseline   & 0.780$\pm$0.003 & \textbf{0.545$\pm$0.002} & & 0.716$\pm$0.006 & \textbf{0.513$\pm$0.003} \\
p-greedy   & 0.936$\pm$0.009 & 0.589$\pm$0.018 & & 0.921$\pm$0.008 & 0.589$\pm$0.019 \\
p-of-max   & \textbf{0.953$\pm$0.004} & 0.545$\pm$0.026 & & \textbf{0.939$\pm$0.003} & 0.536$\pm$0.023\\
p-quantile & 0.935$\pm$0.007 & 0.545$\pm$0.010 & & 0.911$\pm$0.009 & 0.526$\pm$0.006 \\
\bottomrule
\end{tabular}
\end{sc}
\end{small}
\vskip -0.1in
\end{table}

\subsection{Trying other QGFN variants at inference}
\label{sec:extra-variants}
\begin{figure}[h]
\vskip 0.2in
\begin{center}
\centerline{\includegraphics[width=0.5\columnwidth]{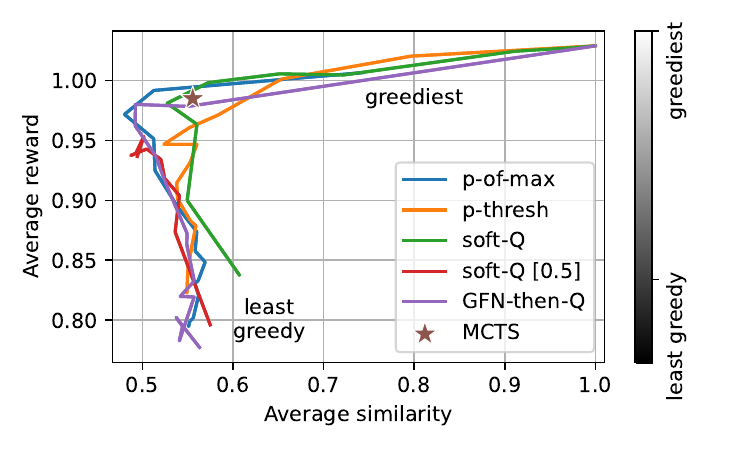}}
\caption{Trying other possible QGFN variants.}
\label{fig:extra_variants}
\end{center}
\vskip -0.2in
\end{figure}

Since the only cost to trying to different ``inference'' variants of QGFN is to code them, we do so out of curiosity. We show the reward/diversity trade-off curves of these variants in Fig.~\ref{fig:extra_variants}, and include $p$-of-max as a baseline variant. As in Fig.~\ref{fig:inference-p} we take 512 samples for each point in the curves (except for MCTS which is more expensive). We try the following:

\begin{itemize}
    \item $p$-thresh, mask all actions where $Q(s,a)<p$;
    \item soft-Q, not really QGFN, but as a baseline simply taking softmax$(Q/T)$ for some temperature $T$, which is varied as the greediness parameter;
    \item soft-Q [0.5], as above but mixed with $P_F$ with a factor $p=0.5$ (i.e. $p$-greedy, but instead of being greedy, use the soft-$Q$ policy);
    \item GFN-then-Q, for the first $Np$ steps, sample from $P_F$, then sample greedily (where $N$ is the maximum trajectory length);
    \item MCTS, a Monte Carlo Tree Search where $P_F$ is used as the expansion prior and $\max_a Q(s,a)$ as the value of a state. Since this is a different sampling method, we run MCTS for a comparable amount of time to other variants, getting about 350 samples, and report the average reward and diversity.
\end{itemize}

We note that these are all tried using parameters from a pretrained $p$-greedy QGFN model. It may be possible for these variants to be much better at inference if the $Q$ used corresponded to the sampling policy.

\subsection{QGFN variants with different objective}

\begin{figure}[h]
\vskip 0.2in
\begin{center}
\centerline{\includegraphics[width=\columnwidth]{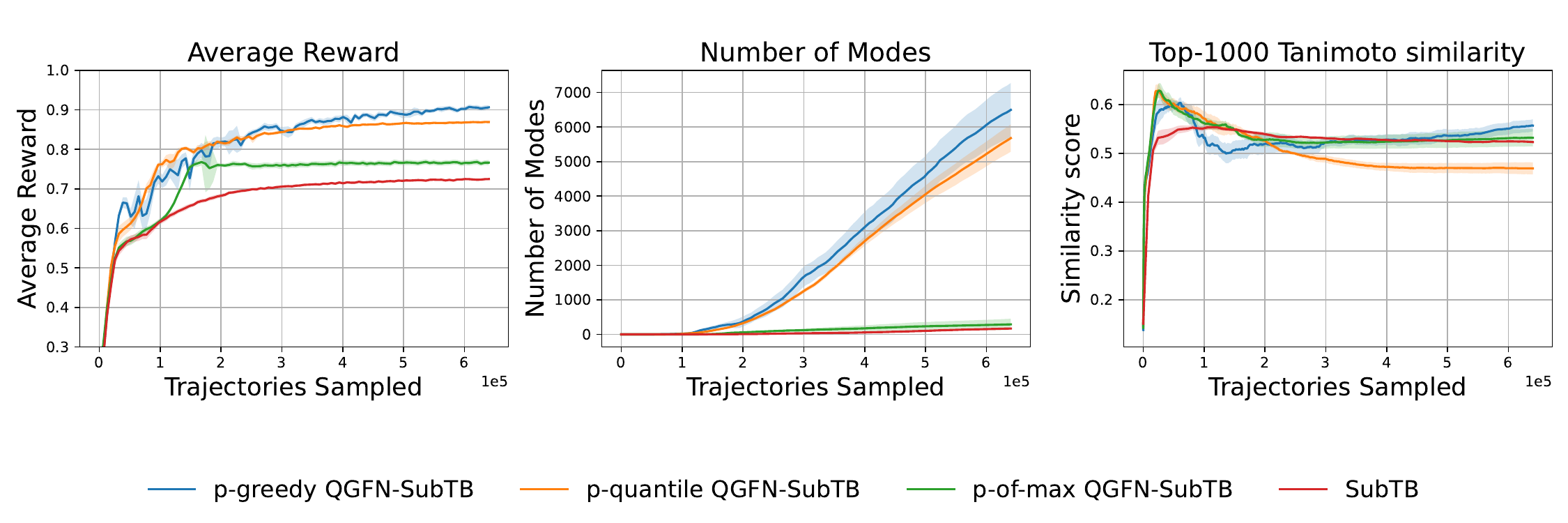}}
\caption{QGFN variants on learning objectives SubTB on Fragment-based molecule task}
\label{fig:results_on_QsubTB}
\end{center}
\vskip -0.2in
\end{figure}

\begin{figure}[h]
\vskip 0.2in
\begin{center}
\centerline{\includegraphics[width=\columnwidth]{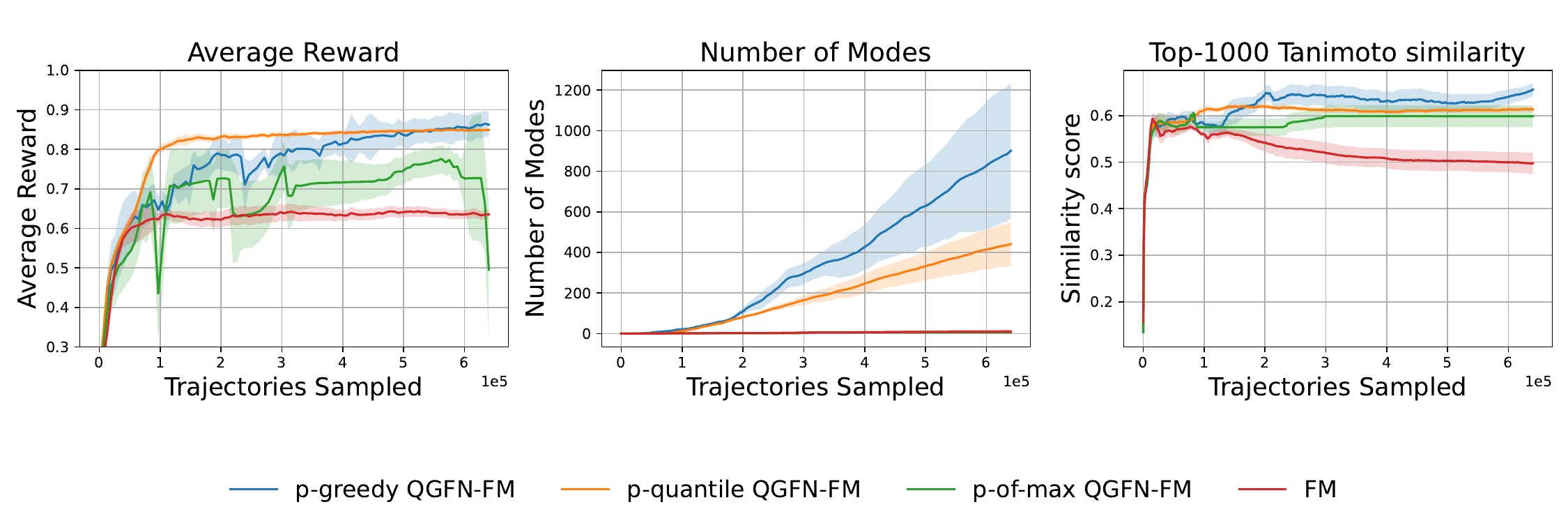}}
\caption{QGFN variants on learning objectives FM (Flow Matching) on Fragment-based molecule task}
\label{fig:results_on_QFM}
\end{center}
\vskip -0.2in
\end{figure}

To demonstrate the robustness of QGFN variants, we explore QGFN with different learning objectives such as SubTB in addition to the TB objective used throughout our experiments. We use the same hyperparameters, except the $p$ values (p-greedy 0.4, p-of-max 0.7, p-quantile 0.7), listed in Table \ref{tab:merged_common_gat} and run the experiments on fragment-based molecule generation. The results are shown in Figure \ref{fig:results_on_QsubTB} and Figure \ref{fig:results_on_QFM}. 

\subsection{Exploring weight sharing in QGFN}
\label{sec:weight-sharing}

\begin{figure*}[!ht]
\vskip 0.2in
\begin{center}
\centerline{\includegraphics[width=\textwidth]{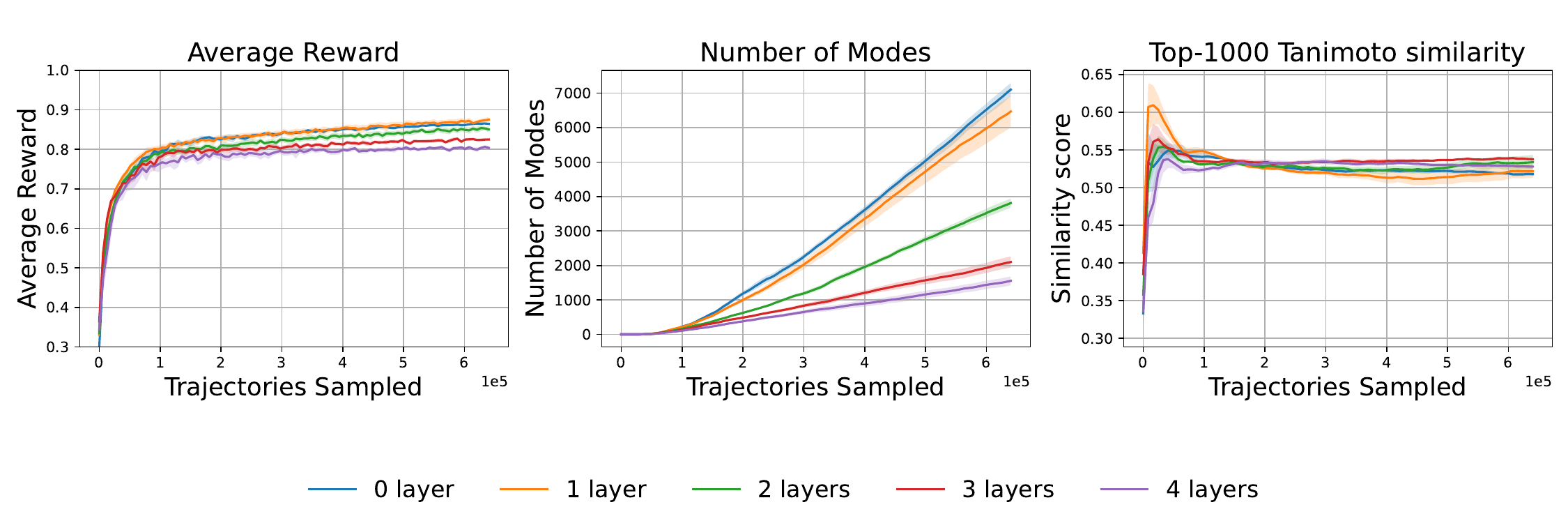}}
\caption{Weight sharing in $p$-greedy QGFN (p=0.4) with different layers}
\label{fig:weight_sharing}
\end{center}
\vskip -0.2in
\end{figure*}

We explore the impact of weight sharing between $P_F$ and $Q$, as they learn from the same environments and training samples. This sharing learning approach could improve in efficiency and performance. Fig.~\ref{fig:weight_sharing} shows the impact of weight sharing on $p$-greedy QGFN, specifically focusing on sharing parameters across various layers of the graph attention transformer in the fragment-based task. 

Unfortunately, naively summing the GFlowNet loss and the $Q$-Learning loss does not yield any improvements, and instead slows down learning. This may be due to several factors; most likely, interference between $Q$ and $P_F$, and a very different scales of the loss function may induce undesirable during training. A natural next stop would be to adjust the relative magnitude of the gradients coming from each loss, and to consider different hyperparameters (perhaps a higher model capacity is necessary), but we leave this to future work. Further exploration in this area could provide additional insights and potentially reduce training complexity. 

\subsection{Training Time and Inference Time Comparison}

In addition to performance, we investigate the training time and inference time for TB and $p$-greedy QGFN. All experiments for this comparison were conducted on NVIDIA V100 GPUs. The results are reported in Table~\ref{tab:training_inference_time}.

\begin{table}[h]
\centering
\caption{Training and inference time comparison between TB and $p$-greedy QGFN.}
\label{tab:training_inference_time}
\begin{tabular}{@{}lcc@{}}
\toprule
 & \textbf{TB} & \textbf{QGFN ($p$-greedy)} \\
\midrule
Training time (10,000 training iterations) & 3 hours, 25 minutes & 6 hours, 20 minutes \\
Inference time (1,000 samples) & 2 min, 44 sec & 2 min, 21 sec \\
\bottomrule
\end{tabular}
\end{table}

\section{Experiments that did not work}
\label{sec:bad-dqgfn}

Several approaches were attempted prior to developing QGFN. These approaches involved sampling from both GFN and DQN independently and learning from the shared data. The key strategies explored include:
\begin{itemize}
\item \textbf{Diverse-Based Replay Buffer}: This method stores batches of trajectories in the replay buffer and samples them based on a pairwise Tanimoto similarity threshold. It aims to diversify the experience replay during training.
\item \textbf{Adaptive Reward Prioritized Replay Buffer}: This strategy stores batches of trajectories in the replay buffer based on the rewards of the samples. In addition, we dynamically adjusts the sampling proportion between GFN and DQN based on the reward performance of the trajectories.
\item \textbf{Weight Sharing}: This involves sharing weights between GFN and DQN to potentially enhance the learning and convergence of the models.
\item \textbf{Pretrained-Q for Greedier Actions}: This method uses a pretrained DQN model for sampling trajectories, helping the GFN to be biased towards greedier actions in the early learning stages.
\item \textbf{$n$-step returns}: As per QGFN, using more than 1-step temporal differences can accelerate temporal credit assignment. This on its own is not enough to solve the tasks used in this work.
\end{itemize}

\begin{figure}[h]
\vskip 0.2in
\begin{center}
\centerline{\includegraphics[width=0.6\columnwidth]{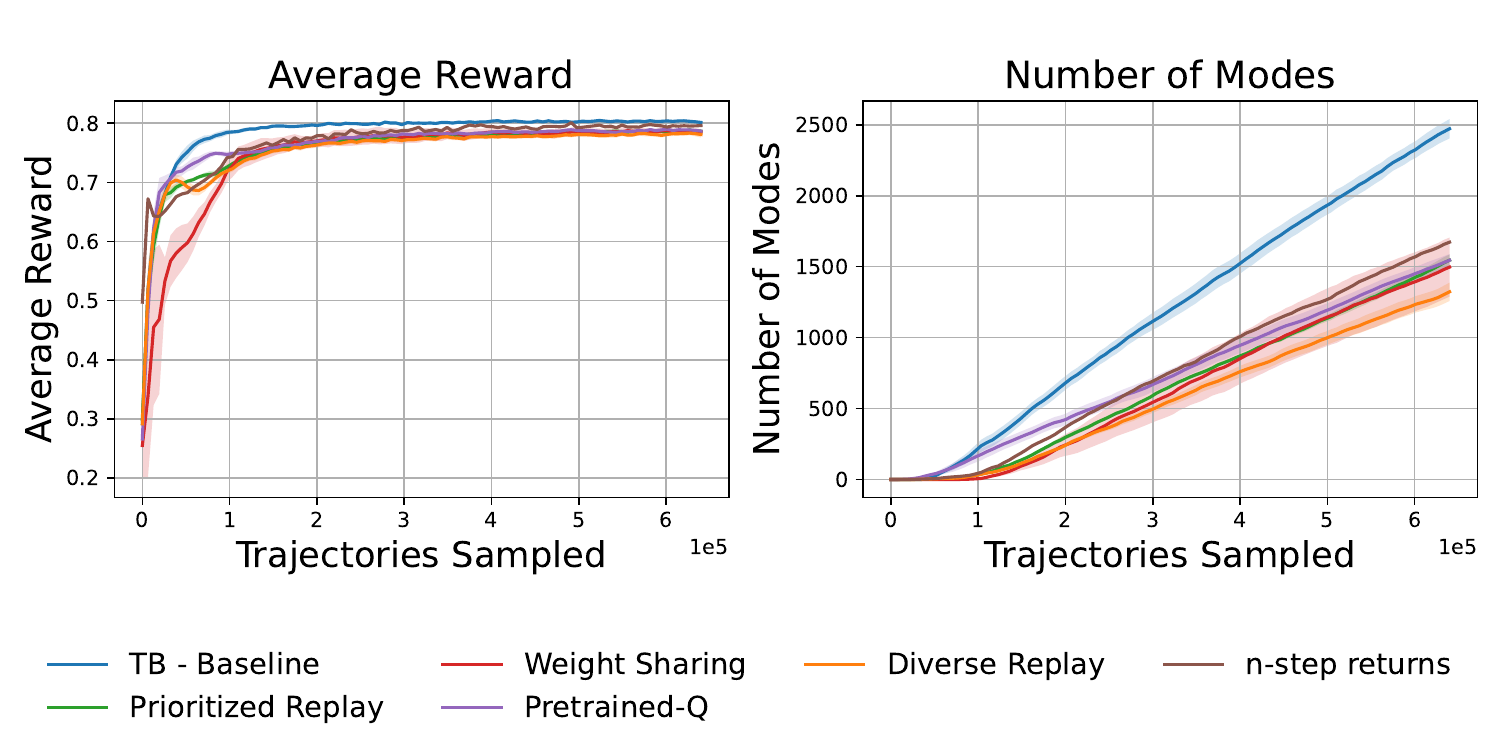}}
\caption{Fragment-based molecule generation task; we showcase the performance of QGFN's predecessor, which failed to beat baselines regardless of our attempts to improve it.}
\label{fig:failed_experiment}
\end{center}
\vskip -0.2in
\end{figure}

Fig \ref{fig:failed_experiment} shows the performance of these approaches evaluated on fragment based molecule generation task. 

\begin{figure*}[h]
\vskip 0.2in
\begin{center}
\centerline{\includegraphics[width=0.9\textwidth]{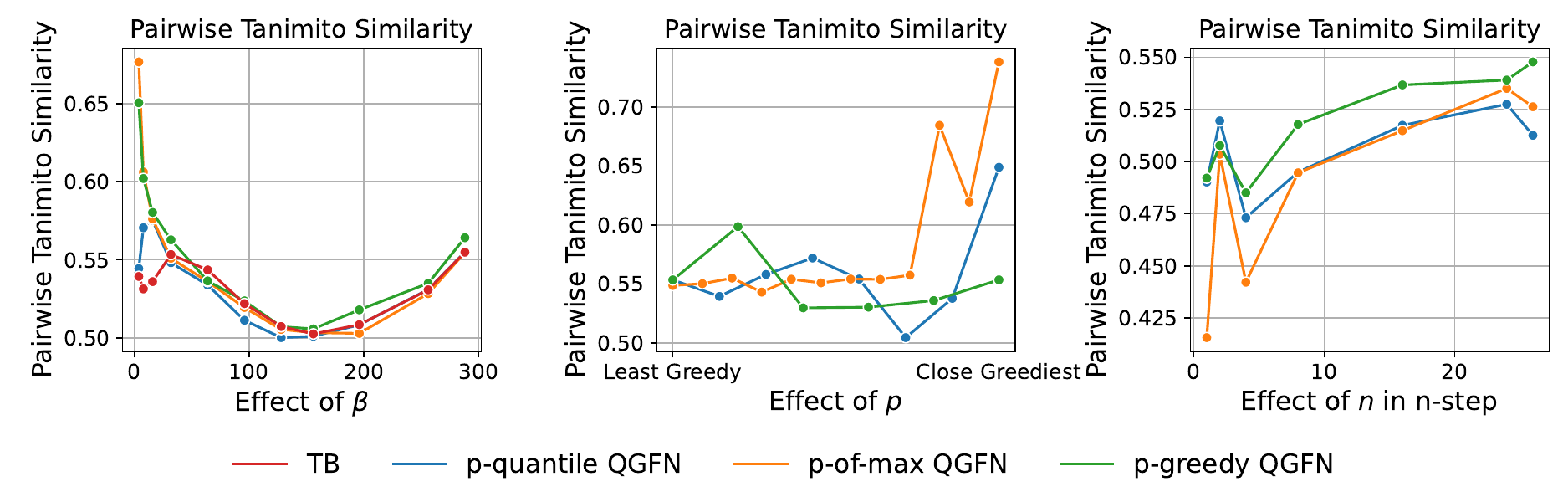}}
\caption{Pairwise Tanimoto Similarity scores assessing the impacts of $\beta$, training parameters $p$ and $n$ in the fragment task. \textit{Left:} An increase in $\beta$ initially decreases sample similarities, followed by a gradual increase in similarity as the models get greedier through $\beta$. \textit{Center:} Increase greediness does not always correlate with sample similarity trade-offs with QGFN, but at peak greediness, similarity scores rebound. \textit{Right:} Increasing $n$ increases similarity among models.}
\label{fig:diversity_ablation}
\end{center}
\vskip -0.2in
\end{figure*}

\newpage 

\section{Full Algorithm}

In this section, we show the detailed implementation of the QGFN algorithm with different variants in Algorithm \ref{alg:QGFN}. For inference, the trained models $P_F$ and $Q$ can be loaded to sample trajectories.

\begin{algorithm*}[]
\caption{QGFN: Full training algorithm details}
 \begin{algorithmic}[1]
 \REQUIRE Reward function $R: \mathcal{X} \rightarrow \mathbb{R}_{>0}$, batch size $M$, Initialize models $P_F$ with parameters $\theta$, $Q(s,a)$ with parameters $\theta'$, greediness parameter $p \in [0, 1]$, training iterations $I$ \\
 \STATE \textbf{For $p$-greedy QGFN:}
 \begin{equation}
    \mu(s'|s) = (1-p)P_F(s'|s) + p\mathbb{I}[s' = \arg\max_i Q(s, i)]
 \end{equation}
 \STATE \textbf{For $p$-quantile QGFN:}
 \begin{equation}
    \mu(s'|s) \propto P_F(s'|s) \mathbb{I}[Q(s, s') \geq q_p(Q, s)]
 \end{equation}
 where $q_p(Q, s)$ is the $p$-quantile of $Q$ over actions at state $s$.
 \STATE \textbf{For $p$-of-max QGFN:}
 \begin{equation}
    \mu(s'|s) \propto P_F(s'|s) \mathbb{I}[Q(s, s') \geq p \max_i Q(s, i)]
 \end{equation}
\FOR{for training iteration $i$ in $I$}

     \FOR{each new trajectory $\tau_j$ from $\tau_1$ to $\tau_M$}{
     \STATE Start $\tau_j$ at $s_0$
     \WHILE{$s_t$ is not terminal}{
        \STATE Sample $s_{t+1}$ from $\mu(s_{t+1}|s_t)$ based on current policy
        \STATE Update $t \gets t+1$
     }\ENDWHILE
     }\ENDFOR

 \STATE Compute trajectory balance loss for $P_F$: $\sum_j\mathcal{L}_{\text{TB}}(\tau_j)$
 \STATE Compute MSE n-step loss for Q-network:
        \[
        \mathcal{L}_{\text{Q}} = \mathbb{E}_{(s_t, a_t)} \left[ \left( Q(s_t, a_t) - G_t^{(n)} \right)^2 \right]
        \]
        where the n-step return $G_t^{(n)}$ is defined as:
        \[
        G_t^{(n)} = \sum_{k=0}^{n-1} \gamma^{k} r_{t+k} + \gamma^{n} \max_{a'} Q(s_{t+n}, a')
        \]
 \STATE Update $\theta$ using $\nabla_\theta \mathcal{L}_{\text{TB}}$;
 \STATE Update $\theta'$ using $\nabla_{\theta'} \mathcal{L}_{\text{Q}}$;
 \ENDFOR
\end{algorithmic}
\label{alg:QGFN}
\end{algorithm*}

\newpage

\section{Experiment details: Fragment-based molecule generation}
\label{sec:experiment_details}

In this section, we give the hyperparameters used for each of our experiments in Tables \ref{tab:merged_common_gat}, and Table \ref{tab:model_specific_params}.  

\begin{table}[H]
\centering
\begin{tabular}{ll}
\toprule
\textbf{Parameter} & \textbf{Value} \\
\midrule
Batch size & 64 \\
Number of steps & 10,000 \\
Optimizer & Adam \\
Number of Layers & 4 \\
Hidden Dim. Size & 128 \\
Number of Heads & 2 \\
Positional Embeddings & Rotary \\
Reward scaling $\beta$ in $R^\beta$ & 32 \\
Learning rate & $1 \times 10^{-4}$ \\
$Z$ Learning rate & $1 \times 10^{-3}$ \\
\bottomrule \\
\end{tabular}
\caption{Hyperparameters and specifications of the Graph Attention Transformer used across all models in Fragment-based molecule generation.}
\label{tab:merged_common_gat}
\end{table}

\begin{table}[H]
\centering
\begin{tabular}{ll}
\toprule
\textbf{Parameter} & \textbf{Value} \\
\midrule
Objective function & TB \\
$p$-greedy & 0.4 \\
$p$-quantile & 0.8 \\
$p$-of-max & 0.91 \\
cosine scheduler for $p$ & 1500 steps \\
Model architecture & Graph Attention Transformer \\
$n$-step & 25 \\
dqn $\tau$ & 0.95 \\
\bottomrule \\
\end{tabular}
\caption{Model-specific parameters for QGFN in Fragment-based molecule generation.}
\label{tab:model_specific_params}
\end{table}

In Table 1, the $p$ values for $p$-greedy, $p$-of-max, and $p$-quantile QGFN are set to 0.4, 0.9858, and 0.93, respectively. These values are selected based on Figure~\ref{fig:inference-p}. The $p$ for $p$-of-max is chosen from \texttt{np.linspace(0.9, 0.999, 16)}, with 0.9858 being one of these values. Similarly, for $p$-quantile, 0.93 corresponds to the second-to-last value from \texttt{np.linspace(0, 1, 16)}. Meanwhile, the 0.4 for $p$-greedy is selected from \texttt{np.linspace(0, 1, 11)}. For LSL-GFN, the chosen $\beta$ is 78, selected from \texttt{np.linspace(64, 128, 65)}.

In our experimental setup, we follow the exact environment specifications and implementations detailed in~\citet{malkin2022trajectory} with the proxy model, used for evaluating molecules, provided by \citet{bengio2021gflownet}. The architecture of the GFlowNet models is based on a graph attention transformer~\citep{velivckovic2017graph}. We set a reward threshold of 0.97 to define modes, with a pairwise Tanimoto similarity criterion of less than $0.65$. RDKit~\citep{2013rdkit} is used to compare pairwise Tanimoto similarity.

To follow closely the original implementation of LSL-GFN described in~\citet{kim2023learning}, we use $\beta \sim U^{[1, 64]}$, where $U$ denotes a uniform distribution. Additionally, we define a simple Multi-layer Perceptron with a hidden size of 256 as the learnable scalar-to-scalar function for the LSL-GFN. For A2C, we use a learning rate of $1\times 10^{-4}$, a training epsilon of $1\times 10^{-2}$, and an entropy regularization coefficient of $1\times 10^{-3}$.
For our SubTB baseline we use SubTB(1), i.e. all trajectories are averaged with equal weight. 

To maintain consistency, the graph attention transformer was used as the model for MunDQN. We sampled 64 trajectories and stored them as transitions in a prioritized replay buffer of size 1,000,000. We then sampled 4096 transitions from the replay buffer to calculate the loss. The Munchausen parameters of 0.10 is selected from $\{0.10, 0.15\}$, an $l_0$ of -2500 and a soft update of $\tau = 0.1$ is used in our experiments. All other parameters are same as the original MunDQN paper in \citet{tiapkin2023generative}.

We also ran an SAC~\citep{haarnoja2018soft} baseline with different $\alpha$ values of 0.5, 0.7, 0.2, along with autotuning, and a $\gamma$ value of 0.99, but we were unable to get it to discover more than 50 modes for the same amount of training iterations and mini-batch sizes.

\subsection{QGFN hyperparameters:}
For all variants of QGFN, we employed a grid-search approach for hyperparameter tuning, with a focus on the parameters $p$ and $n$. Similarly, graph attention transformer is initialized as the $Q$. In the $p$-greedy QGFN variant, we selected a value of $0.4$ for $p$ from the set $\{0.2, 0.4, 0.6, 0.8\}$, and chose an $n$ of 25 from the set $\{1, 2, 4, 8, 16, 24, 25, 26\}$. For the $p$-of-max QGFN variant, a $p$ of 0.91 was chosen from $\{0.2, 0.4, 0.6, 0.8, 0.9, 0.91, 0.93, 0.95\}$, with $n$ again set at 25. To ensure stability throughout the training process, we applied cosine annealing with a single-period cosine schedule over 1500 steps. An additional threshold parameter of $1 \times 10^{-5}$ was applied to the condition $p \max_i Q(s, i) > \text{threshold}$, to prevent the initial training phase from masking actions with very small values. We also introduced a clipping of the Q-value to a minimum of 0 to prevent instability during initial training. For the $p$-quantile QGFN, a $p$ of 0.8 was selected from $\{0.6, 0.7, 0.8, 0.9, 0.95\}$, with $n = 25$. For all variants of QGFN, the DQN employed had a $\tau$ of 0.95, and random action probability of $\epsilon$ was set to 0.1.

\section{Experiment details: QM9} 

\begin{table}[H]
\centering
\begin{tabular}{ll}
\toprule
\textbf{Parameter} & \textbf{Value} \\
\midrule
Batch size & 64 \\
Number of steps & 5,000 \\
Optimizer & Adam \\
Number of Layers & 4 \\
Hidden Dim. Size & 128 \\
Number of Heads & 2 \\
Positional Embeddings & Rotary \\
Reward scaling $\beta$ in $R^\beta$ & 32 \\
Learning rate & $1 \times 10^{-4}$ \\
$Z$ Learning rate & $1 \times 10^{-3}$ \\
\bottomrule \\
\end{tabular}
\caption{Hyperparameters and specifications of the Graph Attention Transformer used across all models in QM9.}
\label{tab:merged_common_gat_qm9}
\end{table}

\begin{table}[H]
\centering
\begin{tabular}{ll}
\toprule
\textbf{Parameter} & \textbf{Value} \\
\midrule
Objective function & TB \\
$p$-greedy & 0.4 \\
$p$-quantile & 0.6 \\
$p$-of-max & 0.6 \\
cosine scheduler for $p$ & 1500 steps \\
Model architecture & Graph Attention Transformer \\
$n$-step & 29 \\
dqn $\tau$ & 0.95 \\
\bottomrule \\
\end{tabular}
\caption{Model-specific parameters for QGFN in QM9.}
\label{tab:model_specific_params_qm9}
\end{table}

In this experiment, we follow the setup described by \citet{jain2023multi}, but only use the HOMO-LUMO gap as a reward signal. The rewards are normalized to fall between [0, 1], although the gap proxy may range from [1,2]. As mentioned in Section \ref{sec:results}, the modes are computed with a reward threshold of 1.10 and a pairwise Tanimoto similarity threshold of 0.70. We employ the same architecture for all models as used in the fragment-based experiments. The training models are 5,000 iterations with a mini-batch size of 64, and $\beta$ is set to 32. RDKit~\citep{2013rdkit} is used to compare pairwise Tanimoto similarity. We train A2C with random action probability 0.01 chosen from $\{0.1, 0.01, 0.001\}$ and entropy regularization coefficient 0.001 chosen from $\{0, 0.1, 0.01, 0.001\}$.Similar to the fragment-based molecule task, we initialized the graph attention transformer with the Munchausen parameter \(\alpha\) set to 0.15, a prioritized replay buffer size of 1,000,000, and a soft update coefficient \(\tau = 0.1\). We sample 64 trajectories and store them in the replay buffer, subsequently sampling 4096 transitions from this buffer. We used the other hyperparameters mentioned in the original MunDQN paper \citep{tiapkin2023generative}.

\subsection{QGFN hyperparameters:} For all variants of QGFN, we employed an exhaustive grid-search approach for hyperparameter tuning, focusing on parameters $p$ and $n$. For $p$-greedy QGFN, we selected 0.4 for $p$ from $\{0.2, 0.4, 0.6\}$ and 29 for $n$ from $\{11, 29, 30\}$. For $p$-of-max QGFN, we chose 0.6 for $p$ from $\{0.3, 0.4, 0.5, 0.6, 0.7, 0.8, 0.9\}$ and 29 for $n$ from $\{28, 29, 30\}$. For $p$-qunatile QGFN, we selected 0.6 for $p$ from $\{0.5, 0.6, 0.7, 0.8\}$ and 29 for $n$ from $\{11, 27, 28, 29\}$. Similarly to the fragment task, we implemented an additional threshold of $ 1 \times 10^{-3}$ to $p \max_i Q(s, i) > \text{threshold}$ and clipped Q-values to a minimum of 0 for stability during initial training. Cosine annealing over 1500 steps is used for all variants. 

\section{Experiment details: RNA-binding task} 
\begin{table}[H]
\centering
\begin{tabular}{ll}
\toprule
\textbf{Parameter} & \textbf{Value} \\
\midrule
Batch size & 64 \\
Number of steps & 5,000 \\
Optimizer & Adam \\
Number of Layers & 4 \\
Hidden Dim. Size & 64 \\
Number of Heads & 2 \\
Positional Embeddings & Rotary \\
Reward scaling $\beta$ in $R^\beta$ & 8 \\
Learning rate & $1 \times 10^{-4}$ \\
$Z$ Learning rate & $1 \times 10^{-2}$ \\
\bottomrule \\
\end{tabular}
\caption{Hyperparameters and specifications of the Sequence Transformer used across all models in RNA-binding task.}
\label{tab:merged_common_gat_rna}
\end{table}

\begin{table}[H]
\centering
\begin{tabular}{ll}
\toprule
\textbf{Parameter} & \textbf{Value} \\
\midrule
Objective function & TB \\
$p$-greedy & 0.4 \\
$p$-quantile & 0.25 \\
$p$-of-max & 0.9 \\
stepwise scheduler for $p$ & 500 steps \\
Model architecture & Sequence Transformer \\
$n$-step & 13 \\
dqn $\tau$ & 0.95 \\
\bottomrule \\
\end{tabular}
\caption{Model-specific parameters for QGFN in RNA-binding task.}
\label{tab:model_specific_params_rna}
\end{table}

We follow the setup of \citet{jain2022biological} but using the task introduced in \citet{sinai2020adalead}. We use a sequence transformer~\citep{vaswani2017attention} architecture with 4 layers, 64-dimensional embeddings, and 2 attention heads. The training for this task is 5000 iterations over mini-batch sizes of 64. The reward scaling parameter $\beta$ is set to 8 and a learning rate of $1 \times 10^{-4}$ and $1 \times 10^{-2}$ for $\log Z$. In this task, $\beta \sim U^{[1, 16]}$ is used for LSL-GFN. Following the approach described by \citet{sinai2020adalead}, the modes are predefined from enumerating the entire RNA landscape for L14RNA1 and L14RNA1+2 to identify local optimal through exhaustive search. ViennaRNA \citep{lorenz2011viennarna} is used to provide the RNA binding landscape. For MunDQN, Munchausen parameter \(\alpha\) set to 0.15, a prioritized replay buffer size of 800,000, and a soft update coefficient \(\tau = 0.1\). We sample 16 trajectories and store them in the replay buffer, subsequently sampling 1024 transitions from this buffer. We used the other hyperparameters mentioned in the original MunDQN paper \citep{tiapkin2023generative}.

\subsection{QGFN hyperparameters:} 
For all QGFN variants, we used grid-search for tuning hyperparameters $p$ and $n$. We used $n = 13$ from the set ${12, 13, 14}$. For $p$-greedy QGFN, $p = 0.4$ was chosen from $\{0.2, 0.4, 0.6, 0.8\}$. For $p$-of-max QGFN, $p = 0.9$ was selected from $\{0.6, 0.7, 0.8, 0.9\}$. In $p$-quantile QGFN, we tried $p = 0.25$ and $0.50$, but neither performed well due to small action spaces. A stepwise scheduler set at 500 steps is applied to $p$-of-max QGFN. 

\section{Experiment details: Bit sequence generation}

\begin{figure}[H]
\vskip -0.2in
\begin{center}
\centerline{\includegraphics[width=0.5\columnwidth]{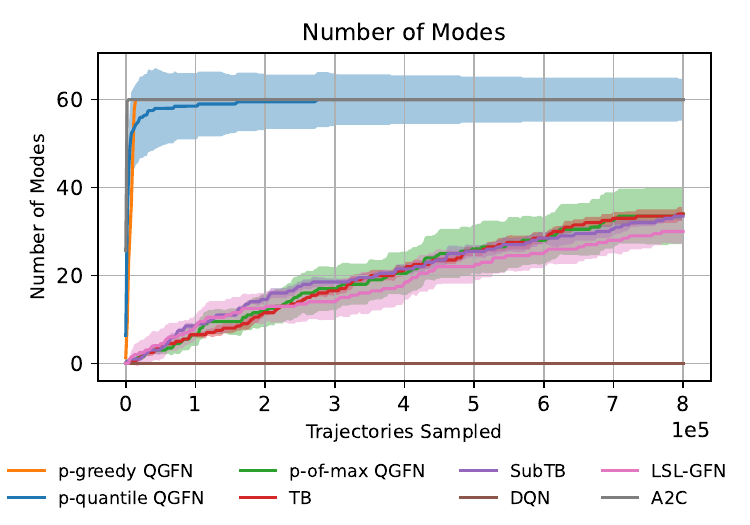}}
\caption{Bit sequence generation, $k = 4$.}
\label{fig:bit-results-4}
\end{center}
\vskip -0.2in
\end{figure}

\begin{table}[H]
\centering
\begin{tabular}{ll}
\toprule
\textbf{Parameter} & \textbf{Value} \\
\midrule
Batch size & 16 \\
Number of steps & 50,000 \\
Optimizer & Adam \\
Number of Layers & 3 \\
Hidden Dim. Size & 64 \\
Number of Heads & 2 \\
Positional Embeddings & Rotary \\
Reward scaling $\beta$ in $R^\beta$ & 3 \\
Learning rate & $1 \times 10^{-4}$ \\
$Z$ Learning rate & $1 \times 10^{-2}$ \\
\bottomrule \\
\end{tabular}
\caption{Hyperparameters and specifications of the Sequence Transformer used across all models in bit sequence generation.}
\label{tab:merged_common_gat_bitseq}
\end{table}

\begin{table}[H]
\centering
\begin{tabular}{ll}
\toprule
\textbf{Parameter} & \textbf{Value} \\
\midrule
Objective function & TB \\
$p$-greedy & 0.4 \\
$p$-quantile & 0.25 \\
$p$-of-max & 0.3 \\
stepwise scheduler for $p$ & 500 steps \\
Model architecture & Sequence Transformer \\
$n$-step & 120 \\
dqn $\tau$ & 0.95 \\
\bottomrule \\
\end{tabular}
\caption{Model-specific parameters for QGFN in bit sequence generation.}
\label{tab:model_specific_params_bitseq}
\end{table}
The bit sequence generation task follows the same environmental setup as \citet{malkin2022trajectory} with $\beta$ value as 3. We generate $|M|=60$ reference sequences by randomly combining $m=15$ symbols from an initial vocabulary $H = \{00000000,11111111,11110000,00001111,00111100\}$.

Beyond the original auto-regressive generative framework in \citet{malkin2022trajectory}, we generate sequences in a prepend-apppend fashion motivated by \citet{shen2023understanding}. We use a sequence transformer \citep{vaswani2017attention} architecture for all experiments with rotary position embeddings, 3 hidden layers with hidden dimension 64 across 8 attention heads. All methods were trained for 50,000 iterations with a minibatch size of 16. For trajectory balance, we use a learning rate of of $1\times 10^{-4}$ for the policy parameters and $1\times 10^{-3}$ for $\log Z$. For SubTB, we use the same hyperparameters as in \citet{madan2023learning}. For LSL-GFN, we use $\beta \sim U^{[1, 6]}$, where $U$ denotes a uniform distribution.

\subsection{QGFN hyperparameters:}
For all variants of QGFN, we did a grid-search approach for hyperparameter tuning for $p$ and $n$. For k=1 where actions are limited to \{0,1\}, we set $n$ at 120, selected from  $\{30, 60, 90, 120\}$, across alla QGFN variants. For the p-greedy QGFN, we chose $p$ = 0.4; for the $p$-of-max QGFN, $p$ was set to 0.3, with cosine annealing applied at 500 steps. In the $p$-qunatile QGFN, we tested $p$ values of 0.25 and 0.50, but neither achieved good performance due to the binary nature of the action space. 

\section{Experiment details: Graph combinatorial optimization problems - maximum independent set (MIS)}

As an additional task, we explore graph combinatorial optimization, specifically the maximum independent set (MIS) problem mentioned in \citet{zhang2023let}. We directly used the codebase shared by \citet{zhang2023let} and report the performance at test time in Table~\ref{tab:MIS_results}.

\begin{table}[H]
\centering
\begin{tabular}{lcc}
\toprule
\textbf{Method} & \textbf{Small - Metric Size} & \textbf{Small - Top 20} \\
\midrule
FL                        & 18.20 & 18.72 \\
FL - QGFN (p-greedy)      & 18.21 & \textbf{19.06} \\
FL - QGFN (p-quantile)         & 18.20 & 18.75 \\
FL - QGFN (p-of-max)        & \textbf{18.26} & 18.74 \\
\bottomrule \\
\end{tabular}
\caption{Comparison of different methods on small graphs: metric size and top 20 metrics.}
\label{tab:MIS_results}
\end{table}

We used the same parameters as in the fragment-based molecule generation experiments. At test time, we applied the p-greedy, p-quantile, and p-of-max sampling strategies respectively for different methods. Note that the reported performance might not reflect the best achievable results on this task, as we did not explore different hyperparameter settings.

\section{Compute Resources}
\label{sec:compute_resources}

All of our experiments were conducted using A100 and V100 GPUs. For the fragment-based task, we used 8 workers on A100 GPUs, and it ran in less than 4 hours. For RNA, we used 4 workers, and it completed in less than 4 hours. For QM9, we used 0 workers, and it finished in less than 9 hours. For Bit sequence, we used 8 workers, and it ran in less than 24 hours.

\section{Broader Impacts}
\label{sec:social_impact}
The research conducted in this work is not far removed from practical applications of generative models. As such, we acknowledge the importance of considering safety and alignment in applications closely related to this work such as drug discovery, material design, and industrial optimization. We believe that research on GFlowNets may lead to models that better generalize, and in the long run may be easier to align. Another important application of GFlowNets is in the causal and reasoning domains; we believe that improving in those fields may lead to easier to understand and safer models.


\newpage
\section*{NeurIPS Paper Checklist}
\begin{enumerate}

\item {\bf Claims}
    \item[] Question: Do the main claims made in the abstract and introduction accurately reflect the paper's contributions and scope?
    \item[] Answer: \answerYes{}
    \item[] Justification: The main claims made in the abstract and introduction are supported by the results presented in Section \ref{sec:results}. Additionally, Section \ref{sec:qgfn-method} provides a thorough analysis of the proposed method, further strengthens the paper's contributions and scope.
    \item[] Guidelines:
    \begin{itemize}
        \item The answer NA means that the abstract and introduction do not include the claims made in the paper.
        \item The abstract and/or introduction should clearly state the claims made, including the contributions made in the paper and important assumptions and limitations. A No or NA answer to this question will not be perceived well by the reviewers. 
        \item The claims made should match theoretical and experimental results, and reflect how much the results can be expected to generalize to other settings. 
        \item It is fine to include aspirational goals as motivation as long as it is clear that these goals are not attained by the paper. 
    \end{itemize}

\item {\bf Limitations}
    \item[] Question: Does the paper discuss the limitations of the work performed by the authors?
    \item[] Answer: \answerYes{}
    \item[] Justification: Section \ref{sec:conclusion} includes a subsection on limitations. Additionally, the discussion on QGFN variants in Section \ref{sec:results} provides a more thorough analysis of the strengths and weaknesses of each variant.

    \item[] Guidelines:
    \begin{itemize}
        \item The answer NA means that the paper has no limitation while the answer No means that the paper has limitations, but those are not discussed in the paper. 
        \item The authors are encouraged to create a separate "Limitations" section in their paper.
        \item The paper should point out any strong assumptions and how robust the results are to violations of these assumptions (e.g., independence assumptions, noiseless settings, model well-specification, asymptotic approximations only holding locally). The authors should reflect on how these assumptions might be violated in practice and what the implications would be.
        \item The authors should reflect on the scope of the claims made, e.g., if the approach was only tested on a few datasets or with a few runs. In general, empirical results often depend on implicit assumptions, which should be articulated.
        \item The authors should reflect on the factors that influence the performance of the approach. For example, a facial recognition algorithm may perform poorly when image resolution is low or images are taken in low lighting. Or a speech-to-text system might not be used reliably to provide closed captions for online lectures because it fails to handle technical jargon.
        \item The authors should discuss the computational efficiency of the proposed algorithms and how they scale with dataset size.
        \item If applicable, the authors should discuss possible limitations of their approach to address problems of privacy and fairness.
        \item While the authors might fear that complete honesty about limitations might be used by reviewers as grounds for rejection, a worse outcome might be that reviewers discover limitations that aren't acknowledged in the paper. The authors should use their best judgment and recognize that individual actions in favor of transparency play an important role in developing norms that preserve the integrity of the community. Reviewers will be specifically instructed to not penalize honesty concerning limitations.
    \end{itemize}

\item {\bf Theory Assumptions and Proofs}
    \item[] Question: For each theoretical result, does the paper provide the full set of assumptions and a complete (and correct) proof?
    \item[] Answer: \answerNA{}
    \item[] Justification: The paper does not include theoretical results
    \item[] Guidelines:
    \begin{itemize}
        \item The answer NA means that the paper does not include theoretical results. 
        \item All the theorems, formulas, and proofs in the paper should be numbered and cross-referenced.
        \item All assumptions should be clearly stated or referenced in the statement of any theorems.
        \item The proofs can either appear in the main paper or the supplemental material, but if they appear in the supplemental material, the authors are encouraged to provide a short proof sketch to provide intuition. 
        \item Inversely, any informal proof provided in the core of the paper should be complemented by formal proofs provided in appendix or supplemental material.
        \item Theorems and Lemmas that the proof relies upon should be properly referenced. 
    \end{itemize}

    \item {\bf Experimental Result Reproducibility}
    \item[] Question: Does the paper fully disclose all the information needed to reproduce the main experimental results of the paper to the extent that it affects the main claims and/or conclusions of the paper (regardless of whether the code and data are provided or not)?
    \item[] Answer: \answerYes{}
    \item[] Justification: In Appendix \S\ref{sec:experiment_details}, we outline all of the hyperparameters needed to reproduce our results, with references to other papers that also used the same environment. In addition, we will be sharing all of our code to run our QGFN algorithm in all tasks presented in this paper.

    \item[] Guidelines:
    \begin{itemize}
        \item The answer NA means that the paper does not include experiments.
        \item If the paper includes experiments, a No answer to this question will not be perceived well by the reviewers: Making the paper reproducible is important, regardless of whether the code and data are provided or not.
        \item If the contribution is a dataset and/or model, the authors should describe the steps taken to make their results reproducible or verifiable. 
        \item Depending on the contribution, reproducibility can be accomplished in various ways. For example, if the contribution is a novel architecture, describing the architecture fully might suffice, or if the contribution is a specific model and empirical evaluation, it may be necessary to either make it possible for others to replicate the model with the same dataset, or provide access to the model. In general. releasing code and data is often one good way to accomplish this, but reproducibility can also be provided via detailed instructions for how to replicate the results, access to a hosted model (e.g., in the case of a large language model), releasing of a model checkpoint, or other means that are appropriate to the research performed.
        \item While NeurIPS does not require releasing code, the conference does require all submissions to provide some reasonable avenue for reproducibility, which may depend on the nature of the contribution. For example
        \begin{enumerate}
            \item If the contribution is primarily a new algorithm, the paper should make it clear how to reproduce that algorithm.
            \item If the contribution is primarily a new model architecture, the paper should describe the architecture clearly and fully.
            \item If the contribution is a new model (e.g., a large language model), then there should either be a way to access this model for reproducing the results or a way to reproduce the model (e.g., with an open-source dataset or instructions for how to construct the dataset).
            \item We recognize that reproducibility may be tricky in some cases, in which case authors are welcome to describe the particular way they provide for reproducibility. In the case of closed-source models, it may be that access to the model is limited in some way (e.g., to registered users), but it should be possible for other researchers to have some path to reproducing or verifying the results.
        \end{enumerate}
    \end{itemize}

\item {\bf Open access to data and code}
    \item[] Question: Does the paper provide open access to the data and code, with sufficient instructions to faithfully reproduce the main experimental results, as described in supplemental material?
    \item[] Answer: \answerYes{}
    \item[] Justification: All of the environments are open access with sufficient instructions provided in Appendix \S\ref{sec:experiment_details}.
    \item[] Guidelines:
    \begin{itemize}
        \item The answer NA means that paper does not include experiments requiring code.
        \item Please see the NeurIPS code and data submission guidelines (\url{https://nips.cc/public/guides/CodeSubmissionPolicy}) for more details.
        \item While we encourage the release of code and data, we understand that this might not be possible, so “No” is an acceptable answer. Papers cannot be rejected simply for not including code, unless this is central to the contribution (e.g., for a new open-source benchmark).
        \item The instructions should contain the exact command and environment needed to run to reproduce the results. See the NeurIPS code and data submission guidelines (\url{https://nips.cc/public/guides/CodeSubmissionPolicy}) for more details.
        \item The authors should provide instructions on data access and preparation, including how to access the raw data, preprocessed data, intermediate data, and generated data, etc.
        \item The authors should provide scripts to reproduce all experimental results for the new proposed method and baselines. If only a subset of experiments are reproducible, they should state which ones are omitted from the script and why.
        \item At submission time, to preserve anonymity, the authors should release anonymized versions (if applicable).
        \item Providing as much information as possible in supplemental material (appended to the paper) is recommended, but including URLs to data and code is permitted.
    \end{itemize}

\item {\bf Experimental Setting/Details}
    \item[] Question: Does the paper specify all the training and test details (e.g., data splits, hyperparameters, how they were chosen, type of optimizer, etc.) necessary to understand the results?
    \item[] Answer: \answerYes{}
    \item[] Justification: All of the environments setting/details are provided in Appendix \S\ref{sec:experiment_details}.
    \item[] Guidelines:
    \begin{itemize}
        \item The answer NA means that the paper does not include experiments.
        \item The experimental setting should be presented in the core of the paper to a level of detail that is necessary to appreciate the results and make sense of them.
        \item The full details can be provided either with the code, in appendix, or as supplemental material.
    \end{itemize}

\item {\bf Experiment Statistical Significance}
    \item[] Question: Does the paper report error bars suitably and correctly defined or other appropriate information about the statistical significance of the experiments?
    \item[] Answer: \answerYes{}
    \item[] Justification: All of our experiments are run with 5 seeds, with interquartile mean and standard error calculated over these 5 seeds.
    \item[] Guidelines:
    \begin{itemize}
        \item The answer NA means that the paper does not include experiments.
        \item The authors should answer "Yes" if the results are accompanied by error bars, confidence intervals, or statistical significance tests, at least for the experiments that support the main claims of the paper.
        \item The factors of variability that the error bars are capturing should be clearly stated (for example, train/test split, initialization, random drawing of some parameter, or overall run with given experimental conditions).
        \item The method for calculating the error bars should be explained (closed form formula, call to a library function, bootstrap, etc.)
        \item The assumptions made should be given (e.g., Normally distributed errors).
        \item It should be clear whether the error bar is the standard deviation or the standard error of the mean.
        \item It is OK to report 1-sigma error bars, but one should state it. The authors should preferably report a 2-sigma error bar than state that they have a 96\% CI, if the hypothesis of Normality of errors is not verified.
        \item For asymmetric distributions, the authors should be careful not to show in tables or figures symmetric error bars that would yield results that are out of range (e.g. negative error rates).
        \item If error bars are reported in tables or plots, The authors should explain in the text how they were calculated and reference the corresponding figures or tables in the text.
    \end{itemize}

\item {\bf Experiments Compute Resources}
    \item[] Question: For each experiment, does the paper provide sufficient information on the computer resources (type of compute workers, memory, time of execution) needed to reproduce the experiments?
    \item[] Answer: \answerYes{}
    \item[] Justification: In appendix \S\ref{sec:compute_resources}, we specified the compute resources used to for our experiments.
    \item[] Guidelines:
    \begin{itemize}
        \item The answer NA means that the paper does not include experiments.
        \item The paper should indicate the type of compute workers CPU or GPU, internal cluster, or cloud provider, including relevant memory and storage.
        \item The paper should provide the amount of compute required for each of the individual experimental runs as well as estimate the total compute. 
        \item The paper should disclose whether the full research project required more compute than the experiments reported in the paper (e.g., preliminary or failed experiments that didn't make it into the paper). 
    \end{itemize}
    
\item {\bf Code Of Ethics}
    \item[] Question: Does the research conducted in the paper conform, in every respect, with the NeurIPS Code of Ethics \url{https://neurips.cc/public/EthicsGuidelines}?
    \item[] Answer: \answerYes{}.
    \item[] Justification: We also consider and discussion societal implications of our work in \S\ref{sec:social_impact}.
    \item[] Guidelines:
    \begin{itemize}
        \item The answer NA means that the authors have not reviewed the NeurIPS Code of Ethics.
        \item If the authors answer No, they should explain the special circumstances that require a deviation from the Code of Ethics.
        \item The authors should make sure to preserve anonymity (e.g., if there is a special consideration due to laws or regulations in their jurisdiction).
    \end{itemize}

\item {\bf Broader Impacts}
    \item[] Question: Does the paper discuss both potential positive societal impacts and negative societal impacts of the work performed?
    \item[] Answer: \answerYes{}
    \item[] Justification: We discuss the positive societal impacts in \S \ref{sec:social_impact}. We do not see any negative social impacts of this paper.
    \item[] Guidelines:
    \begin{itemize}
        \item The answer NA means that there is no societal impact of the work performed.
        \item If the authors answer NA or No, they should explain why their work has no societal impact or why the paper does not address societal impact.
        \item Examples of negative societal impacts include potential malicious or unintended uses (e.g., disinformation, generating fake profiles, surveillance), fairness considerations (e.g., deployment of technologies that could make decisions that unfairly impact specific groups), privacy considerations, and security considerations.
        \item The conference expects that many papers will be foundational research and not tied to particular applications, let alone deployments. However, if there is a direct path to any negative applications, the authors should point it out. For example, it is legitimate to point out that an improvement in the quality of generative models could be used to generate deepfakes for disinformation. On the other hand, it is not needed to point out that a generic algorithm for optimizing neural networks could enable people to train models that generate Deepfakes faster.
        \item The authors should consider possible harms that could arise when the technology is being used as intended and functioning correctly, harms that could arise when the technology is being used as intended but gives incorrect results, and harms following from (intentional or unintentional) misuse of the technology.
        \item If there are negative societal impacts, the authors could also discuss possible mitigation strategies (e.g., gated release of models, providing defenses in addition to attacks, mechanisms for monitoring misuse, mechanisms to monitor how a system learns from feedback over time, improving the efficiency and accessibility of ML).
    \end{itemize}
    
\item {\bf Safeguards}
    \item[] Question: Does the paper describe safeguards that have been put in place for responsible release of data or models that have a high risk for misuse (e.g., pretrained language models, image generators, or scraped datasets)?
    \item[] Answer: \answerNA{}
    \item[] Justification: We do not release models that have a high risk for misuse.
    \item[] Guidelines:
    \begin{itemize}
        \item The answer NA means that the paper poses no such risks.
        \item Released models that have a high risk for misuse or dual-use should be released with necessary safeguards to allow for controlled use of the model, for example by requiring that users adhere to usage guidelines or restrictions to access the model or implementing safety filters. 
        \item Datasets that have been scraped from the Internet could pose safety risks. The authors should describe how they avoided releasing unsafe images.
        \item We recognize that providing effective safeguards is challenging, and many papers do not require this, but we encourage authors to take this into account and make a best faith effort.
    \end{itemize}

\item {\bf Licenses for existing assets}
    \item[] Question: Are the creators or original owners of assets (e.g., code, data, models), used in the paper, properly credited and are the license and terms of use explicitly mentioned and properly respected?
    \item[] Answer: \answerYes{}
    \item[] Justification: We have cited the original paper that produced the code package or dataset throughout the paper. 
    \item[] Guidelines:
    \begin{itemize}
        \item The answer NA means that the paper does not use existing assets.
        \item The authors should cite the original paper that produced the code package or dataset.
        \item The authors should state which version of the asset is used and, if possible, include a URL.
        \item The name of the license (e.g., CC-BY 4.0) should be included for each asset.
        \item For scraped data from a particular source (e.g., website), the copyright and terms of service of that source should be provided.
        \item If assets are released, the license, copyright information, and terms of use in the package should be provided. For popular datasets, \url{paperswithcode.com/datasets} has curated licenses for some datasets. Their licensing guide can help determine the license of a dataset.
        \item For existing datasets that are re-packaged, both the original license and the license of the derived asset (if it has changed) should be provided.
        \item If this information is not available online, the authors are encouraged to reach out to the asset's creators.
    \end{itemize}

\item {\bf New Assets}
    \item[] Question: Are new assets introduced in the paper well documented and is the documentation provided alongside the assets?
    \item[] Answer: \answerNA{}
    \item[] Justification: The paper does not release new assets.
    \item[] Guidelines:
    \begin{itemize}
        \item The answer NA means that the paper does not release new assets.
        \item Researchers should communicate the details of the dataset/code/model as part of their submissions via structured templates. This includes details about training, license, limitations, etc. 
        \item The paper should discuss whether and how consent was obtained from people whose asset is used.
        \item At submission time, remember to anonymize your assets (if applicable). You can either create an anonymized URL or include an anonymized zip file.
    \end{itemize}

\item {\bf Crowdsourcing and Research with Human Subjects}
    \item[] Question: For crowdsourcing experiments and research with human subjects, does the paper include the full text of instructions given to participants and screenshots, if applicable, as well as details about compensation (if any)? 
    \item[] Answer: \answerNA{}
    \item[] Justification: The paper does not involve crowdsourcing nor research with human subjects
    \item[] Guidelines:
    \begin{itemize}
        \item The answer NA means that the paper does not involve crowdsourcing nor research with human subjects.
        \item Including this information in the supplemental material is fine, but if the main contribution of the paper involves human subjects, then as much detail as possible should be included in the main paper. 
        \item According to the NeurIPS Code of Ethics, workers involved in data collection, curation, or other labor should be paid at least the minimum wage in the country of the data collector. 
    \end{itemize}

\item {\bf Institutional Review Board (IRB) Approvals or Equivalent for Research with Human Subjects}
    \item[] Question: Does the paper describe potential risks incurred by study participants, whether such risks were disclosed to the subjects, and whether Institutional Review Board (IRB) approvals (or an equivalent approval/review based on the requirements of your country or institution) were obtained?
    \item[] Answer: \answerNA{}
    \item[] Justification: The paper does not involve crowdsourcing nor research with human subjects.
    \item[] Guidelines:
    \begin{itemize}
        \item The answer NA means that the paper does not involve crowdsourcing nor research with human subjects.
        \item Depending on the country in which research is conducted, IRB approval (or equivalent) may be required for any human subjects research. If you obtained IRB approval, you should clearly state this in the paper. 
        \item We recognize that the procedures for this may vary significantly between institutions and locations, and we expect authors to adhere to the NeurIPS Code of Ethics and the guidelines for their institution. 
        \item For initial submissions, do not include any information that would break anonymity (if applicable), such as the institution conducting the review.
    \end{itemize}

\end{enumerate}

\end{document}